\documentclass{article}

\usepackage{microtype,enumitem}
\usepackage{graphicx}
\usepackage{booktabs} 

\def\W{{\mathbf W}}
\def\J{{\mathbf J}}
\def\I{{\mathbf I}}

\usepackage{hyperref}

\usepackage[accepted]{icml2020}

\usepackage{subcaption}

\icmltitlerunning{Associative Memory in Iterated Overparameterized Sigmoid Autoencoders}

\begin{document}

\twocolumn[
\icmltitle{Associative Memory in Iterated Overparameterized Sigmoid Autoencoders}



\icmlsetsymbol{equal}{*}

\begin{icmlauthorlist}
\icmlauthor{Yibo Jiang}{to}
\icmlauthor{Cengiz Pehlevan}{cbs}
\end{icmlauthorlist}

\icmlaffiliation{to}{John A. Paulson School of Engineering and Applied Sciences, Harvard University, Cambridge, MA, USA}
\icmlaffiliation{cbs}{Center for Brain Science, Harvard University, Cambridge, MA, USA}

\icmlcorrespondingauthor{Cengiz Pehlevan}{cpehlevan@seas.harvard.edu}

\icmlkeywords{Autoencoders, Associate Memory, Neural Tangent Kernel}

\vskip 0.3in
]



\printAffiliationsAndNotice{}  

\begin{abstract}
Recent work showed that overparameterized autoencoders can be trained to implement associative memory via iterative maps, when the trained input-output Jacobian of the network has all of its eigenvalue norms strictly below one. Here, we theoretically analyze this phenomenon for sigmoid networks by leveraging recent developments in deep learning theory, especially the correspondence between training neural networks in the infinite-width limit and performing kernel regression with the Neural Tangent Kernel (NTK). We find that overparameterized sigmoid autoencoders can have attractors in the NTK limit for both training with a single example and multiple examples under certain  conditions. In particular, for multiple training examples, we find that the norm of the largest Jacobian eigenvalue drops below one with increasing input norm, leading to associative memory. 
\end{abstract}

\section{Introduction}
\label{sec:intro}

The mechanisms behind memory have been a long interest of neuroscientists. Hopfield's seminal work proposed that associative memory can be implemented by attractor neural dynamics \cite{hopfield1982neural}, and has been the dominant model that shapes thinking in this domain \cite{hertz2018introduction}. Recently, \citet{radhakrishnan2019overparameterized, radhakrishnan2018memorization} proposed an alternative mechanism and showed that overparameterized autoencoders trained with gradient descent could also implement associative memory in an iterative fashion.  These networks are reported to be easy to train and to not suffer from spurious attractors, unlike Hopfield networks \cite{amit1989associative,hertz2018introduction}. 
The potential benefits of this approach make a theoretical account of it necessary, which we aim to provide here. 



We study auto-encoders in a limit of neural networks that makes theoretical analysis possible. Specifically, as the width of the hidden layers of a feedforward neural network is taken to infinity with a particular initialization scheme, its training dynamics simplifies and can be described by ridgeless kernel interpolation with a kernel called the Neural Tangent Kernel (NTK) \cite{jacot2018neural}. Working in the NTK limit, we examine the input-output Jacobian matrices of the trained networks as they control the stability of trained fixed points. We focus on networks with $\mathrm{sigmoid}$ activation functions, but our results can be extended to other sigmoidal functions such as $\mathrm{erf}$ and $\mathrm{tanh}$ using similar techniques. We make a distinction between the cases of a  single training example and multiple training examples, which exhibit different memory behaviors.

Our main contributions and results are summarized below: 
\begin{itemize}[leftmargin=*,noitemsep,topsep=0pt]
\item First, we analyze autoencoders in the NTK limit trained on a single training example. We argue that the trained Jacobian will stay close to initialization under certain  conditions. 
Therefore, if the initial Jacobian has all eigenvalue norms smaller than 1, this training example will be an \textbf{attractor}. 
\item  Next, we specialize to 2-layer networks. We show that when the norms of training examples are small, attractor formation can \textbf{fail} due to the presence of eigenvalue 1 in the spectrum of trained Jacobian matrices. 
\item We show that a 2-layer network can have \textbf{attractors} when the norms of training examples are large as the induced NTK relies more on the non-linear, saturated region of the activation function. This suggests that the network transitions from a regime where attractor formation fails to a regime where it succeeds as the input norm grows.
\item We verify the predictions of our theoretical results in simulations.
\end{itemize}

Many previous works on generalization \cite{allen2019learning, cao2019generalization} and training \cite{jacot2018neural, du2018gradient, allen2018convergence} in the NTK limit focus on input data on the unit sphere which is violated in practice \cite{krizhevsky2012imagenet}. We highlight how the input norm can set trained neural networks in different learning regions with respect to the trained input-output Jacobian.

Associative memory behavior induced by training overparameterized autoencoders could provide insights into the implicit bias and generalization of neural networks. Autoencoders are trained to learn identity maps, however the existence of attractors indicates failure to learn such maps and thus failure to generalize. Recently, \citet{zhang2019identity} observed that fully connected networks tend to learn a constant function, a global attractor, when trained on a single example. Our single training example results explain this behavior. However, we also show that  attractor formation is dependent on input norm when multiple examples are present, demonstrating that training with a single example is not sufficient to explain the implicit bias of neural networks.

\section{Related Work}
\label{sec:related-work}
Our results use ideas related to NTK and signal propagation in deep networks with random weights. Attractor behavior can also be associated with the implicit bias of deep learning. We review relevant literature from these domains. 

\textbf{Neural Tangent Kernel:} We first review literature on neural networks optimization, especially the NTK theory which we will use extensively through this paper.
Training of neural networks poses a challenging non-convex optimization problem. 
Analysis simplifies if focused on the linearized training dynamics of gradient flow using the NTK theory \cite{jacot2018neural}. The basic idea is that, if initialized properly, in the infinite width limit, parameters of the network stay close to initialization \cite{chizat2019lazy}. Thus, NTK stays relatively constant throughout training. Because NTK governs the training dynamics, positive-definite kernel ensures global convergence of optimization.  Subsequent papers expanded this idea to finite network widths, gradient descent or stochastic gradient descent as opposed to gradient flow, and different loss functions for regression and classification \cite{du2018gradient, allen2018convergence, zou2019improved, lee2019wide}. Also related is research pointing that neural networks at initialization in the infinite width limit behave as Gaussian Processes \cite{lee2017deep, matthews2018gaussian}. 

\textbf{Signal Propagation in Deep Networks with Random Weights:}  In a set of ideas that we will make use of later, \citet{poole2016exponential} developed a mean-field formalism to study layer-to-layer propagation of activation variances and covariances in deep networks with random weights. This line of work \cite{poole2016exponential, schoenholz2016deep} identifies a phase transition between ordered and chaotic regime, where nearby input points converge or diverge as they propagate through the layers, induced by different variances of weight and bias initializations.  Using random matrix theories developed for Gram matrices of neural networks \cite{pennington2017nonlinear},  \citet{pennington2017resurrecting, pennington2018emergence} calculate singular value spectra for input-output Jacobians at initialization and identify its relation to ordered/chaotic training regime. These results cannot be applied to our problem, as they are in a different setting assuming a large depth limit such that the variances for all layers are at the fixed points of the layer-to-layer iterative maps. 

\textbf{Generalization and Implicit Bias:} Associative memory behavior of neural networks can provide insight into generalization of deep learning and implicit bias of gradient descent. Here, we review some recent works in this area, focusing on overparameterized networks and NTK. A pair of recent papers \cite{hayou2019mean, xiao2019disentangling} demonstrate how to use the theories developed for signal propagation to understand NTK regression better and thus trainability and generalization of neural networks. Methods derived from statistical physics also give insight  \cite{cohen2019learning,bordelon2020spectrum}. However, there is a gap in terms of generalization between NTK regression and  training neural networks \cite{allen2019can,arora2019exact} (see however \cite{lee2019wide}), which prompts research on generalization in deep learning beyond NTK \cite{allen2019learning, bai2019beyond}. Another line of research on generalization looks at the implicit bias of gradient descent. Gradient descent on logistic regression can lead to the max-margin solution \cite{soudry2018implicit, ji2018risk}, while optimization on mean squared regression has a shortest path solution \cite{oymak2018overparameterized}. 

\section{Preliminaries}
In this section, we set up our notation and review background material.
\label{sec:pre}
\subsection{Neural Networks}
\label{sec:pre-net}
The output of a neural network is defined by $f(\mathbf{x}) = \tilde{\alpha}^{(L)}(\mathbf{x})$, where the functions $\tilde{\alpha}^{(\ell)}(\cdot): \R^{n_0} \to \R^{n_{\ell}}$ (\textit{preactivations}) and $\alpha^{(\ell)}(\cdot): \R^{n_0} \to \R^{n_{\ell}}$ (\textit{activations}) follow the recursive relation: 
\begin{gather*}
    \alpha^{(0)}(\mathbf{x}) \equiv \mathbf{x} \\
    \tilde{\alpha}^{(\ell + 1)}(\mathbf{x}) \equiv \frac{1}{\sqrt{n_{\ell}}} \mathbf{W}^{(\ell)} \alpha^{(0)}(\mathbf{x}), \quad 
    \alpha^{(\ell)}(\mathbf{x}) \equiv \sigma (\tilde{\alpha}^{(\ell)}(\mathbf{x})),
\end{gather*}
where $\sigma$ is an element-wise activation function and weights are initialized by sampling from an i.i.d. standard Gaussian. We are mostly interested in 
\begin{align}
    \mathrm{sigmoid} = \frac{1}{ 1 + e^{-x}}
\end{align}
as the activation function. We drop the bias term for simplicity. We expect our results to be qualitatively the same with a bias term for sigmoid as the behavior is governed by the activation's shape. Throughout the paper, we will use $f_0(\mathbf{x})$ and $f_{\infty}(\mathbf{x})$ to denote neural networks at initialization and training to zero loss respectively. As shown later, the attractor behavior is mostly governed by the shape of the sigmoidal activations.  

\textbf{Inputs:} Given $n$ training points $\{ \mathbf{x}_i \}_1^n$, we define the following two matrices.
\begin{equation*}
\hat{\mathbf{X}} = \begin{pmatrix}
\kern.6em\vline & \dots & \vline\kern.6em \\
\mathbf{x}_1 & \dots & \mathbf{x}_n \\
\kern.6em\vline & \dots & \vline\kern.6em \\
\end{pmatrix}, \;
f(\hat{\mathbf{X}}) = \begin{pmatrix}
\kern.6em\vline & \dots & \vline\kern.6em \\
f(\mathbf{x}_1) & \dots & f(\mathbf{x}_n) \\
\kern.6em\vline & \dots & \vline\kern.6em \\
\end{pmatrix},
\end{equation*}
where $\hat{\mathbf{X}} \in \R^{n_0 \times n}$ is the data matrix and each column of $f(\hat{\mathbf{X}}) \in \R^{n_0 \times n}$ is the output of the corresponding training example. We further assume that all input examples share the same norm (i.e. $\forall i \;  \Vert\mathbf{x}_i\Vert_2 = r $).

\textbf{Jacobian Matrix:} Given the network setup, the input-output Jacobian matrix can be computed as:
\begin{equation*}
    \begin{split}
        \J(\mathbf{x}) = \frac{1}{\sqrt{n_L}}\mathbf{W}^{(L)} \prod_{k=1}^{L}\left( \mathbf{D}^{(k)}\frac{1}{\sqrt{n_{k-1}}}\mathbf{W}^{(k-1)}\right)
    \end{split}
\end{equation*}
where 
\begin{align*}
    \mathbf{D}^{(k)} = \text{diag}(\dot{\sigma}(\tilde{\alpha}^{(k)}(\mathbf{x}))).
\end{align*}
Here $\dot{}$ denotes first derivative,  and $\text{diag}$ takes in a vector and outputs a diagonal matrix with the vector at the diagonal.  We will use $J_0(\mathbf{x})$ and $J_{\infty}(\mathbf{x})$ to denote Jacobian at initialization and training to zero loss respectively.      

\textbf{Autoencoder:} An autoencoder network is trained via gradient flow to optimize the following loss function:
\begin{equation*}
    \argmin_{f} \frac{1}{2n} \sum_{i=1}^n \Vert f(\mathbf{x}_i) - \mathbf{x}_i\Vert_2^2,
\end{equation*}
where $f$ is the network defined above.

\subsection{Neural Tangent Kernel}
\label{sec:pre-ntk}
In the large-width regime, the neural network $f$ can be approximated by a linearization with respect to its parameters $\bm{\theta}$   \cite{lee2019wide}: 
\begin{equation*}
    f(\mathbf{x}; \bm{\theta}_t) \approx f_0(\mathbf{x}) +  \partial_{\bm{\theta}} f(\mathbf{x})\rvert_{\bm{\theta} = \bm{\theta_0}} (\bm{\theta}_t - \bm{\theta}_0),
\end{equation*}
where $\bm{\theta}_0$ and $\bm{\theta}_t$ are vectors of the network parameters at initialization and time $t$. The first term remains unchanged throughout training. Moreover, we can view $\partial_{\bm{\theta}} f(\mathbf{x})\rvert_{\bm{\theta} = \bm{\theta}_0}$ as a feature map in Hilbert space and derive the following matrix kernel,
\begin{equation*}
    (\bm{\Theta}^{(L)}_0 (\hat{\mathbf{x}}, \mathbf{x}))_{dd'} = \biggl< \partial_{\bm{\theta}} f_d(\hat{\mathbf{x}})\rvert_{\bm{\theta} = \bm{\theta}_0}, \partial_{\bm{\theta}} f_{d'}(\mathbf{x})\rvert_{\bm{\theta} = \bm{\theta}_0} \biggr>,
\end{equation*}
where $\langle  \cdot, \cdot \rangle$ denotes inner product. Indeed, in the infinite width limit \cite{jacot2018neural}, $\bm{\Theta}^{(L)}_0$ converges in probability (stochasticity induced by random initialization) to a deterministic limiting kernel, $\bm{\Theta}^{L}_0 (\hat{\mathbf{x}}, \mathbf{x}) \to \Theta^{(L)}_{\infty}(\hat{\mathbf{x}}, \mathbf{x})  {\bf I}_{n_L}$, where $\Theta^{(L)}_{\infty}$ is a scalar kernel and the training dynamics is entirely governed by it.
 
On the other hand, it can also be shown that in the NTK limit ($n_1, ..., n_{L} \to \infty$ sequentially) and when the weights are initialized by sampling from an i.i.d. standard Gaussian distribution, output functions at each layer tend to be i.i.d. centered Gaussian Processes at initialization \cite{lee2017deep}, and the covariance matrix of layer $\ell$ can be defined recursively by
\begin{equation*}
\begin{split}
\Sigma^{(1)}(\hat{\mathbf{x}}, \mathbf{x}) &= \frac{1}{n_0} \hat{\mathbf{x}}^T \mathbf{x}, \\
\Sigma^{(\ell + 1)}(\hat{\mathbf{x}}, \mathbf{x}) &= \Ex_{g \sim \mathcal{N}(0, \Sigma^{(\ell)})} [\sigma(g(\mathbf{x})) \sigma(g(\hat{\mathbf{x}}))].
\end{split}
\end{equation*}
Under the same limit,  $\Theta^{(L)}_{\infty} $ can also be recursively defined \cite{jacot2018neural}:
\begin{equation*}
    \begin{split}
        \Theta^{(1)}_{\infty}(\hat{\mathbf{x}}, \mathbf{x}) &= \Sigma^{(1)}(\hat{\mathbf{x}}, \mathbf{x}), \\
        \Theta^{(\ell + 1)}_{\infty}(\hat{\mathbf{x}}, \mathbf{x}) &=  \Theta^{(\ell)}_{\infty}(\hat{\mathbf{x}}, \mathbf{x}) \dot{\Sigma}^{(\ell + 1)}(\hat{\mathbf{x}}, \mathbf{x}) + \Sigma^{(\ell + 1)}(\hat{\mathbf{x}}, \mathbf{x}),
    \end{split}
\end{equation*}
where
\begin{equation*}
    \dot{\Sigma}^{(\ell+1)}(\hat{\mathbf{x}}, \mathbf{x}) = \Ex_{g \sim \mathcal{N}(0, \Sigma^{(\ell)})} [\dot{\sigma}(g(\mathbf{x})) \dot{\sigma}(g(\hat{\mathbf{x}}))].
\end{equation*}
For gradient flow training with a least squares loss to zero training error, there is a closed form solution for $f_{\infty}(\mathbf{x})$ using NTK in the infinite width limit  \cite{jacot2018neural}. In the case of autoencoder, we get that, 
\begin{equation}
\label{equ:f-reg}
    f_{\infty}(\mathbf{x}) = \left( \hat{\mathbf{X}} - f_0(\hat{\mathbf{X}}) \right) \tilde{\mathbf{K}}^{-1} \mathbf{k}_x + f_0(\mathbf{x}),
\end{equation}
and
\begin{equation}
\label{equ:j-reg}
    \J_{\infty}(\mathbf{x}) = \left( \hat{\mathbf{X}} - f_0(\hat{\mathbf{X}}) \right) \tilde{\mathbf{K}}^{-1} \frac{\partial \mathbf{k}_x}{\partial\mathbf{x} } + \J_0(\mathbf{x}),
\end{equation}
where
\begin{equation*}
\tilde{\mathbf{K}}_{ij} = \Theta_{\infty}^{L}(\mathbf{x}_i, \mathbf{x}_j), 
\quad  (\mathbf{k}_x)_i = \Theta_{\infty}^{L}(\mathbf{x}_i, \mathbf{x}) .
\end{equation*}

\subsection{Iterative Maps, Attractors, Associative Memory and Jacobian}
\label{sec:pre-map}

We define attractors with respect to an iterative map. Notice that an autoencoder $f$ is a map from $\R^{n_0}$ to $\R^{n_0}$. Therefore, we can apply $f$ iteratively to input $\mathbf{x}$. Formally, we define this sequence for any input $\mathbf{x}$, $\{f^k(\mathbf{x}) \}_{k \in \N}$ where $f^k = \underbrace{f(...f(}_{k}(\mathbf{x})))$. 

\begin{defn}
A fixed point $\mathbf{x}^{*}$ of map $f$ ($f(\mathbf{x}^{*}) = \mathbf{x}^{*}$) is an \textit{attractor} if there exists an open neighborhood of $\mathbf{x}^{*}$ such that for any $\mathbf{x}$ in this neighborhood, $\{f^k(\mathbf{x}) \}_{k \in \N}$ converges to $\mathbf{x}^{*}$ as $k \to \infty$. The set of all such points is called \textit{basin of attraction} of $\mathbf{x}^{*}$. 
\end{defn}

Fixed points attractors can be used to implement associative memory \cite{hopfield1982neural}. A memory clue sets the initial condition of the network dynamics, positioning the network state in a basin of attraction, and the actual memory is recapitulated by the attractor dynamics converging to the corresponding fixed point.

There is a well-know condition for a fixed point to be an attractor \cite{rudin1964principles}. 
\begin{proposition}
\label{prop:att-eigen}
A fixed point $\mathbf{x}^{*}$ is an attractor of a differentiable map $f$ if all eigenvalues of the Jacobian of $f$ at $\mathbf{x}^{*}$ are strictly less than 1 in absolute value. 
\end{proposition}

 Therefore, for a point $\mathbf{x}$ to be an attractor, we need two conditions: (1) $f(\mathbf{x}) = \mathbf{x}$ and (2) all the eigenvalues of $J(\mathbf{x})$ have norm strictly smaller than 1. Condition (1) can be justified in the NTK limit  \cite{jacot2018neural, du2018gradient, allen2018convergence} as long as $\tilde{\mathbf{K}}$ is positive definite. This is theoretically true if all data points live on a sphere, the network has non-polynomial Lipschitz activation function and $L \geq 2$ (c.f. Proposition 2 in \cite{jacot2018neural}). In practice, it is easy to achieve $f(\mathbf{x}) \approx \mathbf{x}$ in overparameterized networks. Therefore,  our focus is on the second condition.

\section{Theoretical Results}

In this section, we present our theoretical results about the attractor behavior of iterated overparametrized autoencoders. We focus on two distinct settings: 
\begin{enumerate}[leftmargin=*,noitemsep,topsep=0pt]
    \item In the first setting, we consider a neural network of any depth in the NTK limit with a single training example.
    \item In the second setting, we focus on a two-layer network in the NTK limit and multiple training examples. 
\end{enumerate}

We first give two key results about the Jacobian norm at network initialization that will be used later. We use the operator norm (induced $l_2$-norm) of Jacobian as a proxy to control the eigenvalue, because the norm is given by the largest singular value, which upper bounds the norm of the largest eigenvalue.

\subsection{A Bound on the Norm of the Jacobian at Network Initialization}
\label{sec:init-jac}
In this section, we first show that with high probability, the operator norm of the initial Jacobian for sigmoid networks drops with increasing depth. To prove this, we use a mathematical technique called the $\epsilon$-net argument \cite{tao2012topics} (Theorem~\ref{thm:epilonnet}) valid for any activation function. We then argue that for sigmoid networks the upper bound of the initial Jacobian norm is concentrated around 0.5 for large $n_0$. This explains the formation of attractors when the trained Jacobian norm stays close to initilization (c.f. Section \ref{sec:single}).

We first prove a proposition to be used in the $\epsilon$-net argument. 
Given a unit vector $\mathbf{z}^{(0)}$ and input $\hat{\mathbf{x}}$, we recursively define  $\tilde{\mathbf{z}}^{(L)} \equiv \mathbf{J}(\hat{\mathbf{x}}) \mathbf{z}^{(0)}$:
\begin{equation*}
        \tilde{\mathbf{z}}^{(\ell)} \equiv \frac{1}{\sqrt{n_{(\ell-1)}}}\mathbf{W}^{(\ell-1)} \mathbf{z}^{(\ell - 1)}, \quad \mathbf{z}^{(\ell)} \equiv \mathbf{D}^{(\ell)} \tilde{\mathbf{z}}^{(\ell)}.
\end{equation*}
The following proposition provides the distribution of $\mathbf{z}^{(\ell)}$. 
\begin{restatable}{proposition}{propinitdist}
\label{prop:init-dist}
For a fixed unit vector $\mathbf{z}^{(0)}$, fixed input data $\hat{\mathbf{x}}$ and a network of depth L at random initialization, with a Lipschitz nonlinearity $\sigma$, and in the limit  $n_1, ..., n_{L-1} \to \infty$, $\J(\hat{\mathbf{x}}) \mathbf{z}^{(0)}$ has the following recursion with ${\mathbf{z}}^{(\ell)}_{i} = \hat{z}^{(\ell)}$:
\begin{equation*}
\begin{split}
\hat{z}^{(1)} &= \sigma'(a)b \quad (a,b) \sim \mathcal{N}\left(\mathbf{0}, 
    \begin{bmatrix}
         \frac{\Vert\hat{\mathbf{x}}\Vert^2_2}{n_0}, & \frac{\hat{\mathbf{x}}^T{ \mathbf{z}^{(0)}}}{n_0}\\
         \frac{\hat{\mathbf{x}}^T{ \mathbf{z}^{(0)}}}{n_0}, & \frac{\Vert \mathbf{z}^{(0)}\Vert^2_2}{n_0} \\
    \end{bmatrix}\right),\\
\hat{z}^{(\ell + 1)} &= \sigma'(a)b \\
&\quad (a,b) \sim  \mathcal{N}\left(\mathbf{0}, 
    \begin{bmatrix}
         \Ex[(\hat{\alpha}^{(\ell)})^2], \! \! \! &\Ex[\hat{\alpha}^{(\ell)} \hat{z}^{(\ell)}] \\
         \Ex[\hat{\alpha}^{(\ell)} \hat{z}^{(\ell)}], \! \! \! & \Ex[(\hat{z}^{(\ell)})^2]\\
    \end{bmatrix}\right),\\
\tilde{\mathbf{z}}^{(L)}_i &= \hat{{z}}^{(L)} \sim \mathcal{N}\left(0, \Ex[(\hat{z}^{(L-1)})^2]\right),
\end{split}
\end{equation*}
where 
\begin{equation*}
    \begin{split}
        \hat{\alpha}^{(1)} &= \sigma(a) \quad a \sim \mathcal{N}\left(0, \frac{\Vert\hat{\mathbf{x}}\Vert_2^2}{n_0}\right), \\
        \hat{\alpha}^{(\ell + 1)} &= \sigma(a) \quad a \sim \mathcal{N}\left(0, \Ex[(\hat{\alpha}^{(\ell)})^2]\right). \\
    \end{split}
\end{equation*}
\end{restatable}
\begin{proof}
See Appendix~\ref{append:init-proof}.
\end{proof}

Using Proposition~\ref{prop:init-dist}, we apply the $\epsilon$-net argument \cite{tao2012topics} (Appendix~\ref{append:init-proof}) to obtain a bound on the Jacobian operator norm. 

\begin{restatable}{theorem}{epilonnet}
\label{thm:epilonnet}
For any data point $\mathbf{x}_i$, $i \in [1,..,n]$, with probability at least $1 - O(n)e^{-O(n_0)}$,
\begin{equation*}
    \left\Vert\J(\mathbf{x}_i)\right\Vert_{op} \leq c \sqrt{n_0 \tau}
\end{equation*}
where $c$ is a constant and $$\tau = \sup_{\mathbf{x_i} \in \hat{\mathbf{X}}, \; \Vert\mathbf{z}^{(0)}\Vert_2= 1 }\Ex[(\hat{z}^{(L-1)})^2 | \mathbf{z}^{(0)}, \mathbf{x}_i]$$
\end{restatable}
\begin{proof}
See Appendix~\ref{append:init-proof}.
\end{proof}

This bound on singular values of $\mathbf{J}(\mathbf{x})$ provides  a way to understand how the Jacobian at network initialization changes with respect to activation functions and the number of layers. Specifically, for $\mathrm{sigmoid}$ activation function, we know that $\dot{\sigma}(x) \in (0, \frac{1}{4}]$ and for any arbitrary unit vector $\mathbf{z}^{(0)}$, %
\begin{equation*}
    \Ex[(\hat{z}^{(L-1)})^2 | \mathbf{z}^{(0)}, \mathbf{x}] = \Ex[\sigma'(a)^2b^2] \leq \frac{\Ex[(\hat{z}^{(L-2)})^2 | \mathbf{z}^{(0)}, \mathbf{x}]}{16} .
\end{equation*}
Thus,
\begin{equation*}
    \Ex[(\hat{z}^{(L-1)})^2 | \mathbf{z}^{(0)}, \mathbf{x}] \leq \frac{1}{n_0 16^{L-1}} \Vert\mathbf{z}^{(0)}\Vert_2^2 =  \frac{1}{n_0 16^{L-1}},
\end{equation*}
and with probability at least $1 - O(n)e^{-O(n_0)}$,
\begin{equation}\label{Jfall}
    \left\Vert\J(\mathbf{x}_i)\right\Vert_{op} \leq \frac{c}{4^{L-1}} \quad \forall i \in [n].
\end{equation}
Therefore, with high probability, the initial Jacobian norm decreases with increasing layers. Based on this result, we choose to study a 2-layer $\mathrm{sigmoid}$ network because it would give an upper bound on the initial Jacobian norm. 

Next, we argue that the largest initial Jacobian norm for a two-layer $\mathrm{sigmoid}$ network is concentrated around $\textbf{1/2}$. For a given training point $\hat{\mathbf{x}}$,
\begin{equation*}
     \J(\hat{\mathbf{x}}) = \frac{1}{\sqrt{n_1}} \mathbf{W}^{(1)} \mathbf{D}^{(1)} \frac{1}{\sqrt{n_0}} \mathbf{W}^{(0)}.
\end{equation*}
From Proposition~\ref{prop:init-dist}, we can reach maximum $\Ex[(\hat{z}^{L-1})^2 | \mathbf{z}^{(0)}, \hat{\mathbf{x}}])$ for any fixed $\mathbf{z}^{(0)}$ when $\hat{\mathbf{x}} = \mathbf{0}$ because every gradient of the hidden layer is at max $\frac{1}{4}$ and the covariance between $\mathbf{x}$ and $\mathbf{z}^{(0)}$ is zero, which would informally suggest that the upper bound of initial Jacobian norm is likely achieved at $\hat{\mathbf{x}} = \mathbf{0}$ (Another way to view this comes from Proposition~\ref{prop:init-dist} where for any given unit vector $\mathbf{z}^{(0)}$, the norm of output vector $\tilde{\mathbf{z}}^{(2)}_i$ has a higher mean when $\hat{\mathbf{x}} = \mathbf{0}$). This gives us the simplification:
\begin{equation*}
    \J(\hat{\mathbf{x}}) = \frac{1}{4}\frac{1}{\sqrt{n_0}} \frac{1}{\sqrt{n_1}}  \mathbf{W}^{(1)} \mathbf{W}^{(0)}.
\end{equation*}
Observe that $\mathbf{W} = \frac{1}{\sqrt{n_1}} \mathbf{W}^{(1)} \mathbf{W}^{(0)}$ is a Gaussian random matrix with $n_1 \to \infty$ where each entry is i.i.d. and the largest singular value of $\frac{1}{\sqrt{n_0}}\mathbf{W}$ is concentrated at $2$ for large $n_0$ \citep{Vershynin}. Consequently, the largest initial Jacobian norm for a $\mathrm{sigmoid}$ network is concentrated around $1/2$. 

\subsection{Training a Multilayer Network with a Single Example}
\label{sec:single}

In this section, we consider the special case when there is only one training example, $\mathbf{x}_1$. We show that under certain conditions the Jacobian stays close to initialization, and, combined with the result from the previous section, the trained network can form attractors. 

We start by analyzing the NTK solution. Note that in this case \eqref{equ:f-reg} can be simplified to:
\begin{equation*}
    \begin{split}
    f_{\infty}(\mathbf{x}) &= \frac{\Theta^{L}_{\infty}(\mathbf{x}, {\mathbf{x}}_1)}{\Theta^{L}_{\infty}({\mathbf{x}}_1, {\mathbf{x}}_1)} ({\mathbf{x}}_1 - f_0({\mathbf{x}}_1)) + f_0(\mathbf{x}).
    \end{split}
\end{equation*}
As we describe below, $\Theta_{\infty}^{(L)}$ tends to a constant kernel as $L\to\infty$ for some network initialization \cite{hayou2019mean} and therefore the trained Jacobian equals the initial one. 
 
 To see this, first, we pay close attention to the covariance matrix $\Sigma^{(L)}$, which is the building block of $\Theta_{\infty}^{(L)}$. We define $q^{(\ell)}_{ab} \equiv \Sigma^{\ell}({\mathbf{x}}_a, \mathbf{x}_b)$
and  $c_{ab}^{(\ell)} \equiv q^{(\ell)}_{ab}/\sqrt{q^{(\ell)}_{aa}q^{(\ell)}_{bb}}$. 
It can be shown that  $c^{*} = 1$ is a fixed point of $c^{(\ell)}_{ab}$ as $\ell\to\infty$ \cite{poole2016exponential}. For sigmoidal networks, the stability of $c^{*}$ is governed by 
\begin{equation*}
    \chi_1 \equiv \left.\frac{\partial c^{(\ell)}_{ab}}{\partial c^{(\ell-1)}_{ab}} \right\rvert_{c = 1} = \Ex[\dot{\sigma}(\sqrt{q^{*}}z)^2], \quad z \sim \mathcal{N}(0,1).
\end{equation*}
where $q^{*}$ is what $q^{(\ell)}_{aa}$ converges to. If $\chi_1 < 1$, $c^{*}$ is a stable fixed point, suggesting that all points become equally similar as they progress through layers. A network under such initialization is said to be in the \textit{ordered region} \cite{schoenholz2016deep}. For such networks, $\Theta_{\infty}^{(\ell)}$ converges to a constant kernel with increasing layers \cite{xiao2019disentangling, hayou2019mean}. As $L\to\infty$, $\partial_{\mathbf{x}} \Theta^{(L)}_{\infty}(\mathbf{x}, \mathbf{x}_1)\rvert_{\mathbf{x} = \mathbf{x}_1} \to 0 $
and,
\begin{equation*}
    \J_{\infty}(\mathbf{x}_1) \underset{L\to\infty}{=} \J_{0}(\mathbf{x}_1),
\end{equation*}
as long as $\Theta^{L}_{\infty}({\mathbf{x}}_1, {\mathbf{x}}_1)$ does not converge to $0$ as $L\to\infty$. We note that this argument applies to other weight and bias variance scaling factors at initialization than the special case (1 and 0 respectively) we focus on. 

Specializing to $\mathrm{sigmoid}$ networks, we first observe that they are in the ordered region  because $\dot{\sigma}(x) \in (0, \frac{1}{4}]$, giving an upper bound on $\chi_1 \leq 1/16$.  Further the lower bound of $\Theta^{(L)}_{\infty}({\mathbf{x}}_1, {\mathbf{x}}_1)$ is $1/4$ (Lemma~\ref{lemma:sigmoid-lower-bound} in Appendix~\ref{append:single-proof}). Therefore, we expect the Jacobian to be constant during training in the large depth limit. %
%
In practice, the trained Jacobian stays close to initialization for $2$ - $3$ layer $\mathrm{sigmoid}$ networks as shown in Section \ref{sec:exp-singl-training}.

Our analysis can explain the empirical results of \cite{zhang2019identity} that fully connected networks trained with single example tend to learn constant functions, leading to memorization.
To see this, remember that for a $\mathrm{sigmoid}$ network the norm of the initial Jacobian falls with increasing the number of layers, eq. \eqref{Jfall}. Other sigmoidal activation functions may show the same behavior. In the same limit, the Jacobian remains constant during training, implying $f_{\infty}(\mathbf{x})$ is approximately constant around $\mathbf{x}_1$.

 The large-depth analysis presented in this section does not carry over to multiple training examples. The limiting constant kernel is singular and fails in training the network to zero loss \cite{jacot2018neural}.
 Instead, we will study 2-layer $\mathrm{sigmoid}$ networks for multiple training examples.


\subsection{Training a 2-Layer Network with Multiple Examples: Linear Region}
\label{sec:mutli-linear}
Next, we consider our second setting of a 2-layer network trained with multiple examples. In this section, we argue that for small input norm $r$, a network in the NTK regime behaves like a linear network resulting in a Jacobian with eigenvalue $1$, and thus may not form associative memory. 

\begin{figure}[t]
\begin{subfigure}{0.48\linewidth}
\includegraphics[width=\linewidth]{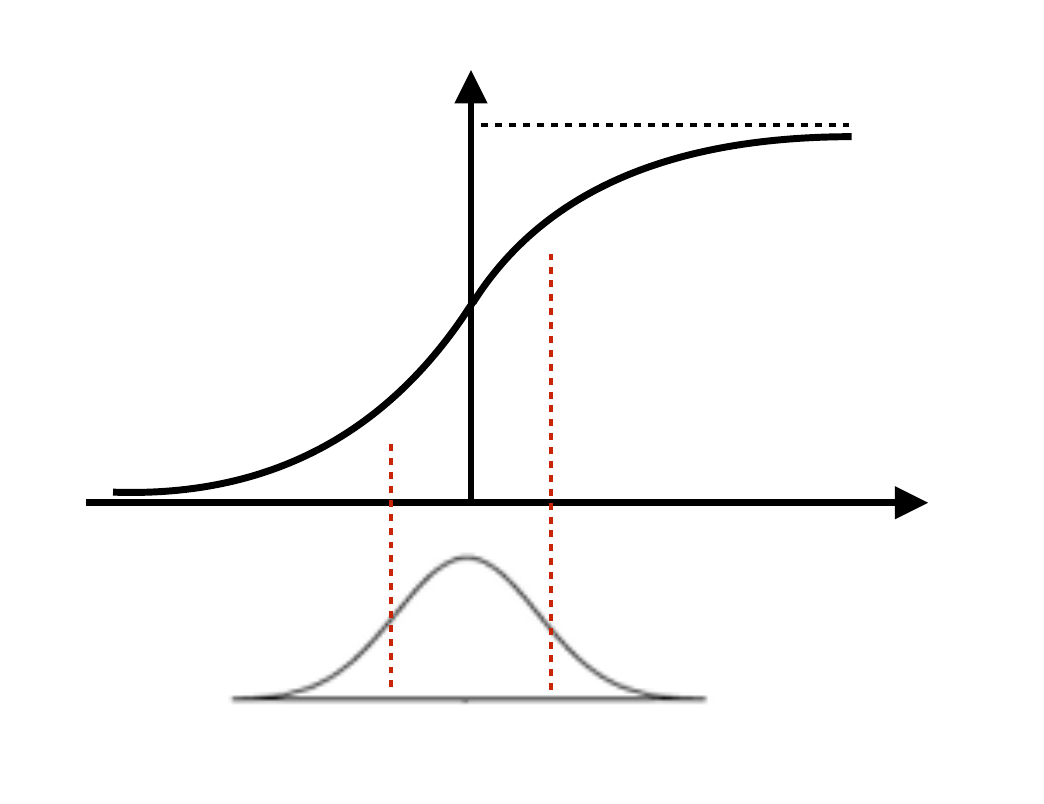}
\centering
\caption{}
\label{fig:linear}
\end{subfigure}
\begin{subfigure}{0.48\linewidth}
\includegraphics[width=\linewidth]{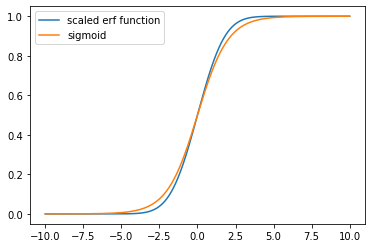}
\centering
\caption{}
\label{fig:sig_erf}
\end{subfigure}
\caption{(a) Linear Region Illustration for Sigmoid: The top graph is a sigmoid function while the bottom one is a Gaussian distribution. The majority of Gasussian distribution falls in the linear region of sigmoid. (b) Sigmoid vs Rescaled Erf.}
\end{figure}

To see this, first note that for a 2-layer network and any two training points $\mathbf{x}_i$ and $\mathbf{x}_j$, NTK can be written as
\begin{equation*}
\Theta^{2}_{\infty}(\mathbf{x}_i, \mathbf{x}_j) =  \Sigma^{1}(\mathbf{x}_i, \mathbf{x}_j) \dot{\Sigma}^{2}(\mathbf{x}_i, \mathbf{x}_j) + \Sigma^{2}(\mathbf{x}_i, \mathbf{x}_j),
\end{equation*}
where $\Sigma^{1}(\mathbf{x}_i, \mathbf{x}_j) = \frac{\mathbf{x}_i^T\mathbf{x}_j}{n_0}$ and both $\Sigma^{2}$ and $\dot{\Sigma}^{2}$ are expectations governed by $\Sigma^{1}$. Because $\Sigma^{1}$ is a covariance matrix defined by inner products, small input norms means small variance for the Gaussian process and the expectation is mostly concentrated to the linear region of the activation. This justifies a linear approximation to activation function. Figure~\ref{fig:linear} shows the $\mathrm{sigmoid}$, which can be approximated linearly around $x=0$ as $\sigma(x) \approx \frac{1}{4}x + \frac{1}{2}$. 

Most of activation functions have similar linear behavior though the range in which this approximation is accurate may differ. Thus, we focus on an arbitrary linear activation function $\alpha x + \beta$ which leads to an initial Jacobian $J_0(\mathbf{x}) = \alpha \frac{1}{\sqrt{n_1}}\frac{1}{\sqrt{n_0}} \mathbf{W}^{(1)}\mathbf{W}^{(0)}$ with initial weights $\mathbf{W}^{(1)}$ and $\mathbf{W}^{(0)}$. We will return to the discussion of $\mathrm{sigmoid}$ at the end of this section. For simplicity, we also assume that $ \hat{\mathbf{X}}$ is full rank and $n_0 \geq n$ as memory tasks tend to have high dimension inputs like images and audios \cite{radhakrishnan2019overparameterized}.

We start examining the Jacobian with the easiest case, $\sigma(x)  = \alpha x$ and $n = n_0$, which leads to  $\J_{\infty}(\mathbf{x}) = \I_{n_0}$.

\begin{restatable}{lemmma}{lemmawarmup}
\label{lemma:warm-up}
Suppose there is a 2-layer network. If the activation function is $\sigma(x) = \alpha x$, $n = n_0$ and the data matrix is full rank. Then at NTK limit, $\J_{\infty}(\mathbf{x}) = \mathbf{I}_{n_0}$.
\end{restatable}
\begin{proof} See Appendix~\ref{append:mutil-linear-proof}.
\end{proof}

In fact, the multiplicity of eigenvalue $1$ is directly related to $n$ under certain conditions.
\begin{restatable}{lemmma}{lemmanobeta}
\label{lemma:no-beta}
Suppose there is a 2-layer network with activation function $\sigma(x) = \alpha x$ and given initial weights $\mathbf{W}^{(1)} \in \R^{n_0 \times n_1}$, $\mathbf{W}^{(0)} \in \R^{n_1 \times n_0}$. If the data matrix is full rank with $n \leq n_0$, then, at the NTK limit ($n_1 \to \infty$), $\J_{\infty}(\mathbf{x})$ has eigenvalue $1$ with multiplicity at least $n$. If at the NTK limit, $\alpha$ is chosen such that $\Vert\J_0(\mathbf{x})\Vert_{op} < 1$, then the multiplicity is exactly $n$ and $1$ is the largest eigenvalue norm.
\end{restatable}
\begin{proof} See Appendix~\ref{append:mutil-linear-proof}.
\end{proof}

\begin{remark}
Lemma~\ref{lemma:no-beta} suggests that a network trained with a single example at convergence can have Jacobian eigenvalue $1$ for $\sigma(x) = \alpha x$ regardless of initial Jacobians. This result may seem to contradict Section~\ref{sec:single}. However, in this case, the argument in Section~\ref{sec:single} is violated because the diagonal value of $\Theta_{\infty}^{(2)}$ is also converging to zero 
as when $\alpha < 1$, the linear activation has a shrinking effect on the layer outputs. 
\end{remark}

The result can be naturally extended to $\sigma(x) = \alpha x + \beta$ with $\beta > 0$.
\begin{restatable}{lemmma}{lemmawithbeta}
\label{lemma:with-beta}
Suppose there is a 2-layer network with activation function $\sigma(x) = \alpha x + \beta$, given initial weights $\mathbf{W}^{(1)} \in \R^{n_0 \times n_1}$, $\mathbf{W}^{(0)} \in \R^{n_1 \times n_0}$ and every data point has the same norm $r$ (i.e. $\forall i \in [n] \;\; \Vert\mathbf{x}\Vert_2 = r $). If the data matrix is full rank with $n \leq n_0$, then, at the NTK limit $n_1 \to \infty$, $\J_{\infty}(\mathbf{x})$ has eigenvalues $1$ with multiplicity at least $n-1$. If at the NTK limit, $\alpha$ and $\beta$ are chosen such that
\begin{equation*}
    \left\Vert \J_0(\mathbf{x})\right\Vert_{op} = 1 - \Delta, \quad \left\Vert\frac{1}{\sqrt{n_1}} \mathbf{W}^{(1)} \mathbf{1}_{n_1}\right\Vert_2 < \frac{\beta n_0 \Delta}{2r\alpha^2},
\end{equation*}
where $0 < \Delta \leq 1$, then the multiplicity is exactly $n-1$ and $1$ is the largest eigenvalue norm.
\end{restatable}
\begin{proof} See Appendix~\ref{append:mutil-linear-proof}.
\end{proof}

Collectively we proved some sufficiency conditions for the largest trained Jacobian eigenvalue to be 1. We emphasize that if the largest eigenvalue is $1$, and not strictly below $1$, the network can fail to form attractors. 

Now, we return to  $\mathrm{sigmoid}$ networks and discuss the implications of Lemma~\ref{lemma:with-beta}. In the linear region, we have $\alpha = \frac{1}{4}$ and $\beta = \frac{1}{2}$. 
As mentioned in Section~\ref{sec:init-jac}, in the NTK limit, $\left\Vert\J_0(\mathbf{x})\right\Vert_2 \approx \frac{1}{2}$, implying $\Delta \approx \frac 12$. On the other hand, $\Vert \frac{1}{\sqrt{n_1}} \mathbf{W}^{(1)} \mathbf{1}_{n_1}\Vert_2$ is the norm of a standard Gaussian vector which follows the chi distribution, and it is concentrated around $\sqrt{n_0}$. 
Then, the conditions in Lemma~\ref{lemma:with-beta} hold with high probability if $r < 2\sqrt{n_0}$. Attractor formation can fail in $\mathrm{sigmoid}$ networks for small input norms, which we observe in simulations (Section \ref{sec:simulations}).


\subsection{Training a 2-Layer Network with Multiple Examples: Beyond Linear Region and a Transition to Attractor Formation}
\label{sec:mutil-beyond-linear}
Section~\ref{sec:mutli-linear} suggests that small $r$ leads to linear behavior and networks may fail to form associative memory. In this section, we explore larger norm inputs where a network in the NTK regime utilizes the non-linear region of the $\mathrm{sigmoid}$. We will argue that in the large norm limit all Jacobian eigenvalues' norms are below 1. Taking into account our result that attractor formation may fail for small values of $r$, we identify a transition with increasing $r$ from a regime where memory formation does not occur to a regime where it occurs. Our results can be adapted for other sigmoidal functions. 

For simplicity, we further assume that there are no parallel inputs in the training data (no $\mathbf{x}$ and $-\mathbf{x}$ at the same time). This is not a hard constraint and an analysis with parallel inputs is given in Appendix~\ref{appnd:parrel-inputs}.

Our strategy is to calculate the Jacobian in the large norm limit using Equation \eqref{equ:j-reg}. We first note for large $r$, 
$\mathbf{X}_{ij} \gg f_0(\mathbf{X})_{ij}$ because all the hidden units in the network is between 0 and 1. Therefore Equation~\ref{equ:j-reg} can be approximated by
\begin{equation}\label{equ:j-reg_app}
\J_{\infty}(\mathbf{x}) \approx  \hat{\mathbf{X}} \tilde{\mathbf{K}}^{-1} \frac{\partial \mathbf{k}_x}{\partial\mathbf{x} } + J_0(\mathbf{x})
\end{equation}

The items of interest here are $\tilde{\mathbf{K}}$ and $\frac{\partial \mathbf{k}_x}{\partial\mathbf{x}}$, for which we need to calculate the NTK and its gradient. We first define an approximation of the NTK and then use it to estimate the Jacobian.

\textbf{Approximation of the NTK:}
%
%
%
Approximating $\mathrm{sigmoid}$ function ($\sigma_s)$ by $\mathrm{erf}$ function there 
$
    \sigma_s(x)  \approx   \frac{1}{2}\text{erf}\left(\frac{1}{2}x\right) + \frac{1}{2}
$ (shown in Figure~\ref{fig:sig_erf}) allows us to use a known closed form solution for NTK \cite{lee2019wide, williams1997computing}.
With this approximation, 2-layer NTK can be written as follows (full derivation can be found in Appendix~\ref{append:deri-ap-ntk}):
\begin{equation}
\begin{split}
\label{eqn:approx}
&{\Theta}^{(2)}_{\infty}(\hat{\mathbf{x}}, \mathbf{x}) \\
&\approx \frac{1}{2\pi} \frac{\hat{\mathbf{x}}^T \mathbf{x}}{\sqrt{(2n_0 + \mathbf{x}^T \mathbf{x})(2n_0 + \hat{\mathbf{x}}^T \hat{\mathbf{x}} ) - (\hat{\mathbf{x}}^T \mathbf{x})^2 )}}  \\
&\,+ \frac{1}{2\pi}\arcsin\bigg( \frac{\hat{\mathbf{x}}^T \mathbf{x}}{\sqrt{( {\mathbf{x}}^T \mathbf{x} + 2n_0)( \hat{\mathbf{x}}^T \hat{\mathbf{x}}+ 2n_0)}}\bigg) + \frac{1}{4}. 
\end{split}
\end{equation}

In order to calculate the gradient of NTK, we define:
\begin{equation*}
    \mathcal{T}(\Sigma, \sigma_1, \sigma_2)(\mathbf{x}_i, \mathbf{x}_j) \equiv \Ex_{f \sim \mathcal{N}(0, \Sigma)} [\sigma_1(f(\mathbf{x}_i))\sigma_2(f(\mathbf{x}_j))]
\end{equation*}
Using this definition, the NTK's gradient for a 2-layer network with arbitrary activation function $\sigma$ is given by, 
\begin{equation*}
\begin{split}
    \frac{\partial \Theta^{(2)}_{\infty}(\hat{\mathbf{x}}, \mathbf{x})}{\partial \mathbf{x}} &=\frac{1}{n_0^2} \hat{\mathbf{x}}^T \mathbf{x} \mathcal{T}(\Sigma^1, \ddot{\sigma}, \ddot{\sigma}) \hat{\mathbf{x}} + \frac{2}{n_0} \mathcal{T}(\Sigma^1, \dot{\sigma}, \dot{\sigma}) \hat{\mathbf{x}}\\
    & +  \frac{1}{n_0^2} \hat{\mathbf{x}}^T \mathbf{x} \mathcal{T}(\Sigma^1, \dot{\sigma}, \dddot{\sigma}) \mathbf{x} + \frac{1}{n_0} \mathcal{T}(\Sigma^1, \sigma, \ddot{\sigma}) \mathbf{x}.
\end{split}
\end{equation*}

Now we can use these expressions to calculate $\tilde{\mathbf{K}}$ and $\frac{\partial \mathbf{k}_x}{\partial\mathbf{x}}$ in Equation~\ref{equ:j-reg_app}.

\textbf{Calculation of $\tilde{\mathbf{K}}$ and $\frac{\partial \mathbf{k}_x}{\partial\mathbf{x}}$: }
\label{sec:kernel-entry}
 We first look at Equation~\eqref{eqn:approx}, using the fact that because all the data points have the same norm $\mathbf{x}_i^T \mathbf{x}_j = r^2\rho_{i,j}$ with $|\rho_{i,j}| < 1$ (since there are no parallel inputs):
\begin{equation*}
\begin{split}
\tilde{\mathbf{K}}_{ij} &  
= \frac{1}{4} + \frac{1}{2\pi}\arcsin\left( \frac{{r^2\rho_{i,j}}}{r^2 + 2n_0}\right)\\
 & \quad + 
 \frac{1}{2\pi} \frac{r^2\rho_{i,j}}{\sqrt{(2n_0 + r^2)^2 - r^4\rho_{i,j}^2}} .
\end{split}
\end{equation*}

To get insight into the behavior of $\tilde{\mathbf{K}}_{ij}$, let's consider diagonal and off-diagonal entries separately in the large-$r$ limit. First the off-diagonals:
\begin{equation*}
\tilde{\mathbf{K}}_{ij}
\approx  \frac{1}{4} + \frac{1}{2\pi}\arcsin\left( \rho_{i,j}\right)+
\frac{1}{2\pi} \frac{\rho_{i,j}}{\sqrt{1 - \rho_{i,j}^2 }}.
\end{equation*}
Next we look at the diagonals:
\begin{equation*}
\begin{split}
\tilde{\mathbf{K}}_{ii}  \approx \frac{r}{4\pi\sqrt{n_0}}.
\end{split}
\end{equation*}
The diagonal grows linearly with $r$ and dominates over the off-diagonals. The kernel tends to a scaled identity  matrix in the $r\to \infty$ limit. 
Thus, 
%
\begin{equation}\label{Km1}
    \left\Vert \tilde{\mathbf{K}}^{-1}\right\Vert_{op} \approx \frac{4\pi\sqrt{ n_0}}{r}.
\end{equation}

Next, we look at $\frac{\partial \mathbf{k}_x}{\partial\mathbf{x}}$. As we are interested in trained Jacobian at training examples, without loss of generality, we focus on $\J_{\infty}(\mathbf{x}_1)$. We already know that $\tilde{\mathbf{K}}$ tends to converge to a diagonal matrix with large $r$. One could use this result to suggest that $\frac{\partial \Theta_{\infty}^{2}(\mathbf{x}_i, \mathbf{x})}{\partial\mathbf{x}}\rvert_{\mathbf{x}_j} \approx 0$ if $i \neq j$. In fact, with large $r$, it can be shown that (Appendix~\ref{app_dkdx}):
 \begin{equation}\label{dkdx}
 \begin{split}
 \left.\frac{\partial \mathbf{k}_x}{\partial\mathbf{x} }\right\rvert_{\mathbf{x}_1} \approx
 \begin{bmatrix}
 \frac{\partial \Theta_{\infty}^{L}(\mathbf{x}_1, \mathbf{x})}{\partial\mathbf{x}}\rvert_{\mathbf{x}_1},
 \mathbf{0},
 \dots,
 \mathbf{0} 
 \end{bmatrix}^T
 \end{split}
 \end{equation}
 and
 \begin{align}\label{dkdx}
 \left\Vert \frac{\partial \mathbf{k}_x}{\partial\mathbf{x}} \right\Vert_{op} \approx \frac{1}{8\pi \sqrt{n_0}} .
 \end{align}

\textbf{Trained Jacobian and attractor formation: }
Finally, we can use our results in Equations \eqref{Km1} and \eqref{dkdx} to calculate the trained Jacobian using Equation \eqref{equ:j-reg_app} with large $r$:
\begin{equation*}
\begin{split}
&\left\Vert \J_{\infty}(\mathbf{x})\right\Vert _{op} \approx \left\Vert \hat{\mathbf{X}} \tilde{\mathbf{K}}^{-1} \frac{\partial \mathbf{k}_x}{\partial\mathbf{x} } + J_0(\mathbf{x})\right\Vert_{op} \\
&\leq \left\Vert \mathbf{X} \tilde{\mathbf{K}}^{-1} \frac{\partial \mathbf{k}_x}{\partial\mathbf{x} }\right\Vert_{op} + \left\Vert \J_0(\mathbf{x})\right\Vert_{op} \\
& \approx r  \frac{4\pi\sqrt{ n_0}}{r}  \frac{1}{8\pi \sqrt{n_0}} + \left\Vert \J_0(\mathbf{x})\right\Vert_{op}\\
&= \frac{1}{2} + \left\Vert \J_0(\mathbf{x})\right\Vert_{op}.
\end{split}
\end{equation*}
To go from the second to the third row of the equation above, we used the fact that each column of $\mathbf{X}$ has norm $r$, and in the limit $\tilde{\mathbf{K}}^{-1}$ is a scaled identity matrix and $\frac{\partial \mathbf{k}_x}{\partial\mathbf{x} }$ has only one non-zero row and that row is a scaled $\mathbf{x}_1$ (Appendix~\ref{app_dkdx}).

To see the implications of this result, we observe that as $r \to \infty$, $\J_0(\mathbf{x}) \approx \mathbf{0}$ because the pre-activation of the two layer network will be infinitely big and the units saturate leading to zero gradients in the Jacobian matrix, we can conclude in this limit
\begin{equation*}
    \left\Vert\J_{\infty}(\mathbf{x})\right\Vert_{op} \leq 1/2.
\end{equation*}
Combined with our previous result that attractor formation may fail for small values of $r$, this suggests that there is a transition from a region where associative memory formation fails to region where it succeeds with increasing $r$. We show  this transition in simulations in Section \ref{sec:simulations} and verify that the largest eigenvalue of the Jacobian falls towards $1/2$ asymptotically. 


\subsection{Beyond NTK}
\label{sec:addition}
In practice, large values of $r$ often require a larger width to stay in NTK. If the large width condition is violated, trained weights will not stay close to initialization. However, in this case, a significant deviation of weights from initialization caused by gradient descent may saturate hidden units, leading to zero gradients in the Jacobian matrix. As shown in Figure~\ref{fig:2-layer-sigmoid-curve-32}, for fixed hidden size, larger input radius leads to near zero eigenvalues when the NTK conditions no longer hold. Therefore, the network can behave as if only the last layer is trained with saturated hidden features and close to zero Jacobian norm since all the $\mathbf{D}^{(\ell)}$ matrices have mostly zero diagonal entries. Note that, in overparameterized networks, this type of optimization can often result in zero training loss as well due to the over-parameterization of the last layer.


\section{Simulations}\label{sec:simulations}

\subsection{Experiment Setup}
\textbf{Training and iterative convergence criteria:} Training is stopped when the training loss of the auto-encoder drops below a threshold, which we chose to be $10^{-7}$. Iterative convergence happens when a non-fixed point converges to a fixed point measured by mean-squared-error after passing through the trained autoencoder iteratively. This threshold was $10^{-2}$.

\textbf{Implementation details:} We used vanilla gradient descent with learning rate $1$, similar to \cite{jacot2018neural}. The code is implemented with Pytorch. For each setting, we ran experiments 100 times to get 100 sets of samples. For all experiments except experiments on MNIST, the samples are randomly generated.

\subsection{Single Training Example}

\begin{figure}[t]
    \centering
    \includegraphics[width = 0.5\linewidth]{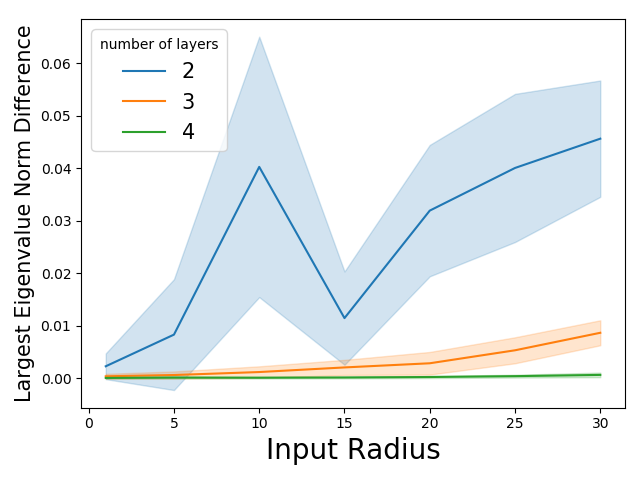}
\caption{Difference of the largest eigenvalue norms  at initialization and after training, i.e. $\abs{\Vert\lambda_1(\J_{0})\Vert  - \Vert\lambda_1(\J_{\infty})\Vert}$.}
\label{fig:2-3-4-layer}
\end{figure}

\label{sec:exp-singl-training}
We first present experiments for a single training example. The $\mathrm{sigmoid}$ network is chosen to have hidden dimension $1000$ with input dimension $32$.  Figure~\ref{fig:2-3-4-layer} shows that the difference between the largest eigenvalue norms at initialization and after training decreases with more number of layers.

\subsection{Multiple Training Examples}

\textbf{Linear Region:} In this section, we first illustrate the eigenvalue distribution in the linear region by sampling unit vectors as data ($r=1$). Here, we trained 2 layer $\mathrm{sigmoid}$ networks with input dimension $10$ and hidden size $1000$ for $2$, $5$ and $8$ training points. As suggested by Lemma~\ref{lemma:with-beta}, there should be $n-1$ eigenvalues with norm around $1$. This is supported by Figure~\ref{fig:2-layer-linear} in the Appendix, where  $10\%$, $40\%$ and $70\%$ of the eigenvalues are near 1. Here, the presence of eigenvalue 1 indicates that the network operates in the linear region.

\begin{figure*}[t]
    \centering
    \begin{subfigure}{0.3\linewidth}
        \centering
        \includegraphics[height=1.2in]{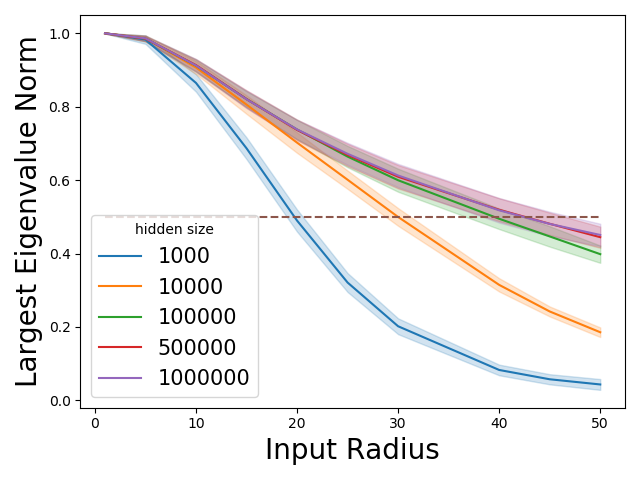}
        \caption{number of training points: 5}
    \end{subfigure}%
    ~ 
    \begin{subfigure}{0.3\linewidth}
        \centering
        \includegraphics[height=1.2in]{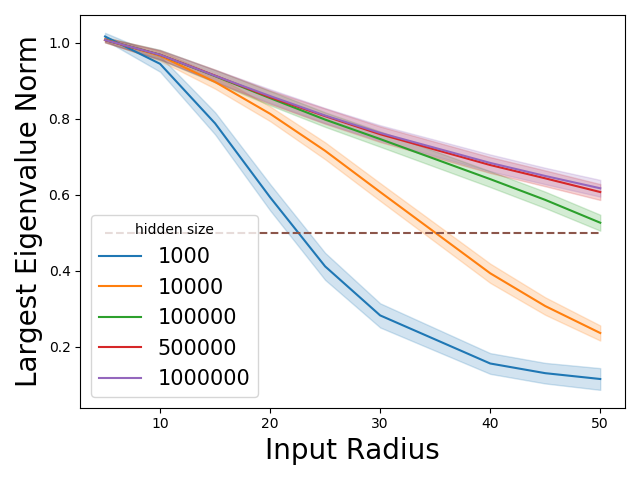}
        \caption{number of training points: 20}
    \end{subfigure}
    ~ 
    \begin{subfigure}{0.3\linewidth}
        \centering
        \includegraphics[height=1.2in]{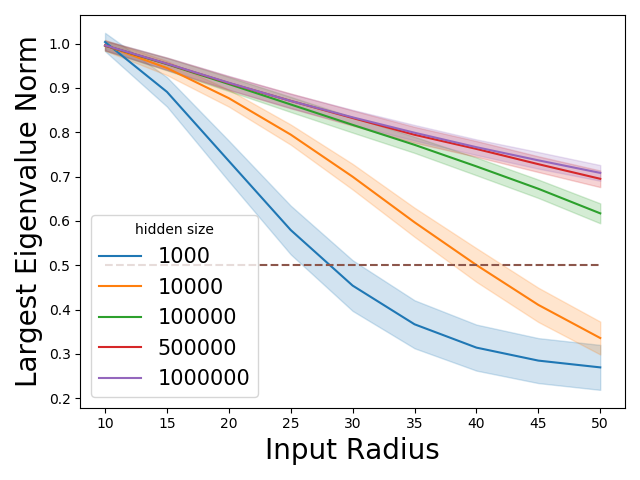}
        \caption{number of training points: 40}
    \end{subfigure}
    \caption{Largest eigenvalue norm vs input norm: input dimension 32. } 
\label{fig:2-layer-sigmoid-curve-32}
\end{figure*}

\textbf{Beyond Linear Region:} We demonstrate how the largest eigenvalue norm varies with the input radius. We test with various hidden dimensions to achieve the NTK limit on input dimension $32$ with the number of training data $5$, $20$, and $40$. Only experiments that can be trained to have loss below $10^{-7}$ or fit into single Titan V GPU are included.

The general trend, as shown in Figure~\ref{fig:2-layer-sigmoid-curve-32}, is that as we move away from the linear region, the largest eigenvalue norm will drop, and as we keep increasing the hidden layer size, it will move close to the $1/2$ limit as suggested by our analysis. It is also worth noting that the input radius needs to be increased with large number of training examples to get training loss below $10^{-7}$ under reasonable iterations, implying the capacity of the networks is controlled by input radius and agreeing with the intuition and theoretical results from \citep{allen2018convergence}, that there needs to be some level of separation between data points to train the network.


\subsection{Basin of Attraction}
\begin{figure}
\centering
\begin{subfigure}{ 0.48\linewidth}
    \includegraphics[width = \linewidth]{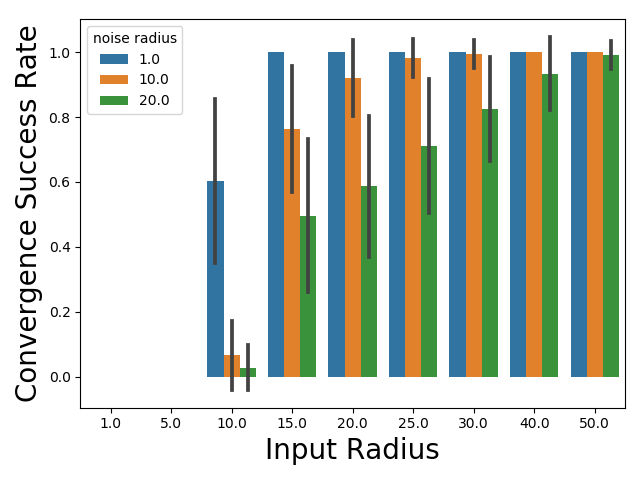}
\caption{Random vectors}
\label{fig:basin-5}
\end{subfigure}
\begin{subfigure}{ 0.48\linewidth}
    \includegraphics[width = 1\linewidth]{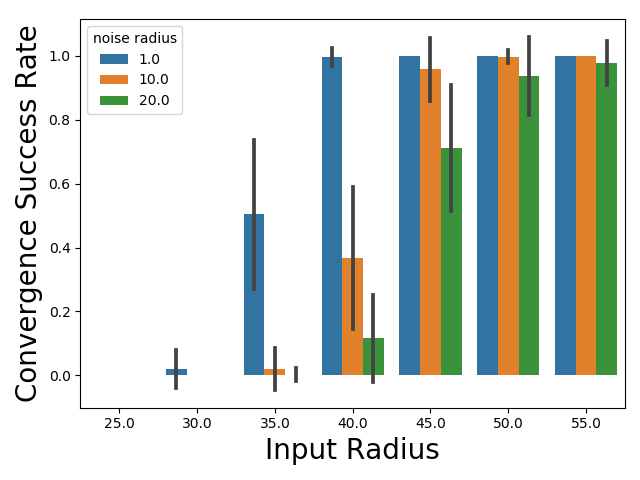}
\caption{MNIST}
\label{fig:basin--mnist-5}
 \end{subfigure}
 \caption{Convergence Success Rate vs Input Radius: 5 training examples}
\end{figure}

We test basin of attraction by adding Gaussian noise to training examples and check if the modified examples can converge to the original ones via iterative maps under 50 iterations. And the convergence rate is the number of samples that could be successfully recovered. The standard deviation of the Gaussian noise is called the noise radius. The network has two layers with hidden size $10000$ and input dimension $32$. Not surprisingly, Figure~\ref{fig:basin-5} shows that larger input norm gives greater basin of attraction. More experiments can be found in Appendix~\ref{append:basin}.

\subsection{MNIST Data}
We also test  basin of attraction experiments on MNIST dataset to check if we can recover real training examples. The images are prepossessed by subtracting means and rescaled to have different input norms for testing. Similar to the setting before, Figure~\ref{fig:basin--mnist-5} also shows that larger input norm gives greater basin of attraction. Notice that because MNIST images have large input dimension, they need larger radius to move out of the linear region. More experiments can be found in Appendix~\ref{append:basin-mnist}. 

\subsection{Sigmoidal Activation}
\begin{figure}
    \centering
    \includegraphics[width = 0.5\linewidth]{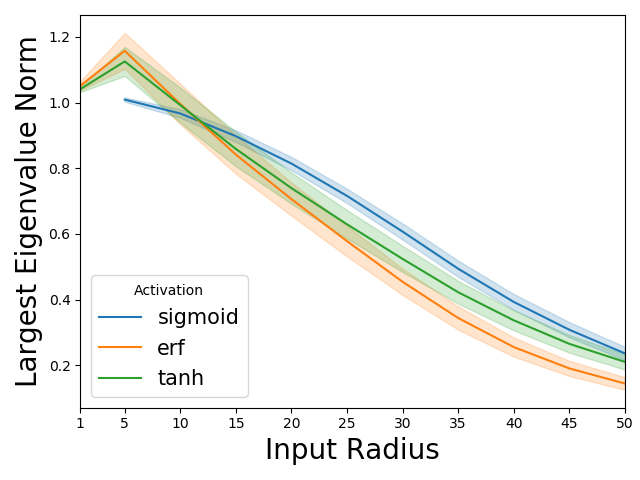}
\caption{Input Radius and Eigenvalue Norm Curve for Different Activation Functions}
\label{fig:act}
\end{figure}

Finally, we show that our results can be extended to different sigmoidal activation functions as well. We chose two-layer network with hidden size $10000$ and input dimension $32$ and $20$ training examples. As before, only settings that led to a training loss below $10^{-7}$ are included. Figure~\ref{fig:act} clearly shows that all the activation functions share similar curves. Notice that both $\mathrm{tanh}$ and $\mathrm{erf}$ have large eigenvalue when $r$ is small. This is not a contradiction to our Lemma~\ref{lemma:no-beta} as their $\alpha = \dot{\sigma}(0)$ is too large to satisfy the conditions in Lemma~\ref{lemma:no-beta}. To further verify this result, we also include how the histogram of eigenvalue norm changes for those activations in the Appendix~\ref{append:act-eigen-histo}.

\section{Conclusion}
In this paper, we theoretically and empirically show that training overparameterized sigmoid autoencoders can lead to attractors for a single training example and multiple training examples in the non-linear region, with the help of theories developed in the NTK limit. We identified a behavior change governed by the input radius. Some future directions include generalizing our results to other activations, identifying other factors that can determine whether autoencoders can learn to have attractors or not, and how the formations of attractors are related to generalization in deep learning.

\section*{Acknowledgements} C. Pehlevan thanks the Harvard Data Science Initiative, Google and Intel for support.

\bibliography{ref}
\bibliographystyle{icml2020}

\clearpage
\onecolumn

\appendix
\counterwithin{figure}{section}
\section{Proofs for Sec~\ref{sec:init-jac}}
\label{append:init-proof}
\propinitdist*
\begin{proof}
We will prove this by induction for $\ell = 1, ..., L-1$.

\textbf{Basic Step}
\begin{equation*}
    \begin{split}
        {\mathbf{z}}^{(1)}_i = \sigma'\left(\frac{1}{\sqrt{n_0}} (\mathbf{W}^{(0)}_i)^T \mathbf{\hat{x}}\right) \frac{1}{\sqrt{n_0}}(\mathbf{W}^{(0)}_i)^T \mathbf{z}^{(0)}
    \end{split}
\end{equation*}
Notice that $\mathbf{W}^{(0)}_i \sim \mathcal{N}(\mathbf{0}, \mathbf{I}_{n_0})$. Thus, we have the following:
\begin{equation*}
    \begin{split}
        a &= \frac{1}{\sqrt{n_0}} (\mathbf{W}^{(0)}_i)^T \mathbf{\hat{x}} \sim \mathcal{N}\left(0, \frac{\Vert\mathbf{\hat{x}}\Vert_2^2}{n_0}\right) \\
        b &= \frac{1}{\sqrt{n_0}} (\mathbf{W}^{(0)}_i)^T \mathbf{z}^{(0)} \sim \mathcal{N}\left(0, \frac{\Vert \mathbf{z}^{(0)}\Vert_2^2}{n_0}\right)
    \end{split}
\end{equation*}
$a$ and $b$ are not independent:
\begin{equation*}
    \Ex[ab] = \Ex[(\frac{1}{\sqrt{n_0}} (\mathbf{W}^{(0)}_i)^T \mathbf{\hat{x}}) (\frac{1}{\sqrt{n_0}} (\mathbf{W}^{(0)}_i)^T \mathbf{z}^{(0)})] = \frac{1}{n_0} \mathbf{\hat{x}}^T \Ex[(\mathbf{W}^{(0)}_i)(\mathbf{W}^{(0)}_i)^T]  \mathbf{z}^{(0)} = \frac{1}{n_0} \mathbf{\hat{x}}^T \I_{n_0}  \mathbf{z}^{(0)} = \frac{\mathbf{\hat{x}}^T  \mathbf{z}^{(0)}}{n_0}
\end{equation*}
Note that the result is independent of the index $i$, we can define $\hat{z}^{(1)} = {\mathbf{z}}^{(1)}_{i}$. Therefore, the base step has been proven. 

\textbf{Inductive Step}
\begin{equation*}
    \begin{split}
        {\mathbf{z}}^{(\ell+1)}_i = \sigma'(\frac{1}{\sqrt{n_{\ell}}} (\mathbf{W}^{(\ell)}_i)^T \tilde{\alpha}^{(\ell)}(\mathbf{\hat{x}})) \frac{1}{\sqrt{n_\ell}}(\mathbf{W}^{(\ell)}_i)^T {\mathbf{z}}^{(\ell)}
    \end{split}
\end{equation*}
Then,
\begin{equation*}
    \begin{split}
        a = \frac{1}{\sqrt{n_{\ell}}} (\mathbf{W}^{(\ell)}_i)^T \tilde{\alpha}^{(\ell)}(\mathbf{\hat{x}})\sim \mathcal{N}(\mathbf{0}, \frac{1}{n_\ell} \sum^{n_\ell}_{i=0} ((\tilde{\alpha}^{(\ell)}(\mathbf{\hat{x}})_i)^2)
    \end{split}
\end{equation*}
With $n_1, ..., n_{\ell} \to \infty$, $\text{Var}(a) = \Ex[(\hat{\alpha}^{(\ell)})^2]$. Similarly, 
\begin{equation*}
\begin{split}
        b &= \frac{1}{\sqrt{n_\ell}}(\mathbf{W}^{(\ell)}_i)^T {\mathbf{z}}^{(\ell)} \\
        b & \sim \mathcal{N}(0, \Ex[(\hat{z}^{(\ell)})^2] ) \quad \text{if } n_1, ..., n_{\ell} \to \infty
\end{split}
\end{equation*}
On the other hand, 
\begin{equation*}
    \begin{split}
        \Ex[ab] &= \Ex[(\frac{1}{\sqrt{n_\ell}} (\mathbf{W}^{(\ell)}_i)^T \tilde{\alpha}^{(\ell)}(\mathbf{x}) ) (\frac{1}{\sqrt{n_\ell}} (\mathbf{W}^{(\ell)}_i)^T {\mathbf{z}}^{(\ell)} )] 
        = \frac{1}{n_\ell} (\tilde{\alpha}^{(\ell)}(\mathbf{x}))^T \Ex[(\mathbf{W}^{(\ell)}_i)(\mathbf{W}^{(\ell)}_i)^T] {\mathbf{z}}^{(\ell)}
         = \frac{1}{n_\ell} (\tilde{\alpha}^{(\ell)}(\mathbf{x}))^T {\mathbf{z}}^{(\ell)}\\
         & = \Ex[\hat{\alpha}^{(\ell)} \hat{{z}}^{(\ell)}] \quad \text{if } n_1, ..., n_{\ell} \to \infty 
    \end{split}
\end{equation*}
The recursive definition is now proven up to layer $\ell -1$. Now let's look at the last layer.
\begin{equation*}
    \tilde{\mathbf{z}}^{(L)}_i = \frac{1}{\sqrt{n_{L-1}}}\mathbf{W}^{(L-1)}_i \mathbf{z}^{(L-1)}
\end{equation*}
By similar arguments as before, it is easy to show that with $n_1, ..., n_{L-1} \to \infty$, $\tilde{\mathbf{z}}^{(L)}_i  \sim \mathcal{N}(0, \Ex[(\hat{z}^{(L-1)})^2])$. This concludes the proof.
\end{proof}

\epilonnet*
\begin{proof}
For a fixed unit vector $\mathbf{z}^{(0)}$ and fixed input $\hat{\mathbf{x}}$, we know that based on Proposition~\ref{prop:init-dist}, $\tilde{\mathbf{z}}^{(L)}_i \sim \mathcal{N}(0, \Ex[(\hat{z}^{(L-1)})^2 | \mathbf{z}^{(0)}, \hat{\mathbf{x}}])$. Define $z$ as
\begin{equation*}
z = \frac{1}{\Ex[(\hat{z}^{(L-1)})^2 | \mathbf{z}^{(0)}, \hat{\mathbf{x}}]}\Vert\tilde{\mathbf{z}}^{(L)}\Vert_2^2 =  \chi^2_{n_0}
\end{equation*}
First, notice that we can have the following tail bound for chi-square distribution (for instance, \citep{kolar2012marginal})
\begin{equation*}
\text{Pr}[|z/n_0 - 1|\geq \epsilon] \leq \exp (-\frac{3}{16}n_0 \epsilon^2)
\end{equation*}
when $\epsilon \in [0, 1/2)$. In this case, let $\epsilon = \frac{1}{3}$. Consider a subset of coordinates $M$ with cardinality $|M| \leq O(n_0)$ \citep{allen2018convergence}. Taking the $\epsilon$ ball $\mathcal{B}$ of this subspace with $\epsilon=1/3$, we know what
\begin{equation*}
    |\mathcal{B}| \leq 7^{|M|} = e^{|M|ln7} = e^{O(n_0)}
\end{equation*}
Then, taking the union bound for all unit vectors in $\mathcal{B}$, we know that
\begin{equation*}
\begin{split}
    \forall \mathbf{z}_0 \in \mathcal{B} \quad
    &\bigcup\limits_{z_0} \text{Pr}[|z/n_0 - 1|\geq \frac{1}{3}] \\
    &\leq \exp (-\frac{1}{48}n_0) \exp(O(n_0)) \leq \exp (-O(n_0))
\end{split}
\end{equation*}
Therefore, by the $\epsilon$-net argument \citep{tao2012topics}, for any unit vector $\mathbf{u}$ with only non-zero entries in $M$, we have with probability $1 - \exp (-O(n_0))$, 
\begin{equation*}
    \Vert\J(\hat{\mathbf{x}})\mathbf{u}\Vert_2^2 \leq 2 n_0 \tau \Vert\mathbf{u}\Vert_2^2 = C^2 \Vert\mathbf{u}\Vert_2^2
\end{equation*}
For any arbitrary vector $\mathbf{v}$, we can decompose it in the following way: $\mathbf{v} = \mathbf{u}_1 + \mathbf{u}_2 + ... + \mathbf{u}_K$ with $K = O(1)$ where each $\mathbf{u}_i$ comes from a different non-overlapping coordinate set $M$. 
\begin{equation*}
\begin{split}
    \Vert\J(\hat{\mathbf{x}})\mathbf{v}\Vert_2 \leq C \sum_{i=1}^{K} \Vert\mathbf{u}_i\Vert_2 &\leq C \sqrt{K} (\sum_{i=1}^K \Vert\mathbf{u}_i\Vert_2^2)^{1/2}\\
    &\leq O(1) C \Vert\mathbf{v}\Vert.
\end{split}
\end{equation*}
Thus, with probability at least $1 - O(1)\exp(-O(n_0))$,
\begin{equation*}
\begin{split}
    \left\Vert \J(\hat{\mathbf{x}})\right\Vert_{op}  \leq O(1) C &= O(1) \sqrt{2 n_0 \tau} \\
    &= c \sqrt{n_0 \tau},
\end{split}
\end{equation*}
where $c$ is a constant. Taking the union bound over all the data points concludes the proof.
\end{proof}

\section{Proofs for Sec~\ref{sec:single}}
\label{append:single-proof}
\begin{restatable}{lemmma}{lowerbound}
\label{lemma:sigmoid-lower-bound}
Under the setting in Section~\ref{sec:pre-net} with $\mathrm{sigmoid}$ as the activation function, 
\begin{equation*}
    \Theta_{\infty}^{(L)}(\mathbf{x}, \mathbf{x}) \geq \frac{1}{4}
\end{equation*}
\end{restatable}
\begin{proof}
For any $\ell$, we have
\begin{equation*}
    \begin{split}
        \Theta_{\infty}^{(\ell + 1)}(\mathbf{x},\mathbf{x}) &= \Theta_{\infty}^{(\ell)}(\mathbf{x},\mathbf{x}) \dot{\Sigma}^{(\ell+1)}(\mathbf{x},\mathbf{x}) + \Sigma^{(\ell+1)}(\mathbf{x},\mathbf{x}) \\
        & \geq \Sigma^{(\ell+1)}(\mathbf{x},\mathbf{x}) \\
        & = \Ex_{g \sim \mathcal{N}(0, \Sigma^{(\ell)})}[\sigma(g(\mathbf{x}))^2] \\
        & =  \Ex_{g \sim \mathcal{N}(0, \Sigma^{(\ell)})}\left[\left(\sigma(g(\mathbf{x})) - \frac{1}{2}\right)^2\right] + \frac{1}{4} \\
        & \geq \frac{1}{4}
    \end{split}
\end{equation*}
where $\sigma(f(\mathbf{x})) - \frac{1}{2}$ moves $\mathrm{sigmoid}$ function to the origin such that it is an odd function. 
\end{proof}

\section{Proofs for Sec~\ref{sec:mutli-linear}}
\subsection{Main Lemmas}
\label{append:mutil-linear-proof}
\lemmawarmup*
\begin{proof}
\begin{equation*}
    \begin{split}
        \J_{\infty}(\mathbf{x}) &= \frac{2\alpha^2}{n_0} (\hat{\mathbf{X}} - f_0(\hat{\mathbf{X}}))(\frac{2\alpha^2}{n_0} \hat{\mathbf{X}}^T \hat{\mathbf{X}})^{-1} \hat{\mathbf{X}}^T + J_{0}(\mathbf{x})\\
    \end{split}
\end{equation*}
Notice that $\J_0(\mathbf{x}) = \alpha \frac{1}{\sqrt{n_1}}\frac{1}{\sqrt{n_0}} \W^{(1)}\W^{(0)}$ and $f_0(\hat{\mathbf{X}}) = \alpha \frac{1}{\sqrt{n_1}}\frac{1}{\sqrt{n_0}} \W^{(1)}\W^{(0)}\hat{\mathbf{X}} = \J_0(x)\hat{\mathbf{X}} $.
\begin{equation*}
\begin{split}
\J_{\infty}(\mathbf{x}) &= \J_0(\mathbf{x}) - f_0(\hat{\mathbf{X}})( \hat{\mathbf{X}}^T \hat{\mathbf{X}})^{-1}\hat{\mathbf{X}}^T + \hat{\mathbf{X}}(\hat{\mathbf{X}}^T \hat{\mathbf{X}})^{-1}\hat{\mathbf{X}}^T \\
&= \alpha \frac{1}{\sqrt{n_1}}\frac{1}{\sqrt{n_0}} \W^{(1)}\W^{(0)} - \alpha \frac{1}{\sqrt{n_1}}\frac{1}{\sqrt{n_0}} \W^{(1)}\W^{(0)}\hat{\mathbf{X}}(\hat{\mathbf{X}}^T \hat{\mathbf{X}})^{-1}\hat{\mathbf{X}}^T + \hat{\mathbf{X}}(\hat{\mathbf{X}}^T \hat{\mathbf{X}})^{-1}\hat{\mathbf{X}}^T\\
&= \alpha \frac{1}{\sqrt{n_1}}\frac{1}{\sqrt{n_0}} \W^{(1)}\W^{(0)} - \alpha \frac{1}{\sqrt{n_1}}\frac{1}{\sqrt{n_0}} \W^{(1)}\W^{(0)}\I_{n_0} + \I_{n_0} \\
& = \I_{n_0}
\end{split}
\end{equation*}
\end{proof}

\lemmanobeta*
\begin{proof}
Based on the proof of last section, we know that
\begin{equation*}
\begin{split}
\J_{\infty}(\mathbf{x}) &= \J_0(\mathbf{x}) - f_0(\hat{\mathbf{X}})( \hat{\mathbf{X}}^T \hat{\mathbf{X}})^{-1}\hat{\mathbf{X}}^T + \hat{\mathbf{X}}(\hat{\mathbf{X}}^T \hat{\mathbf{X}})^{-1}\hat{\mathbf{X}}^T \\
&= \J_0(\mathbf{x}) - \J_0(\mathbf{x})\hat{\mathbf{X}} ( \hat{\mathbf{X}}^T \hat{\mathbf{X}})^{-1}\hat{\mathbf{X}}^T + \hat{\mathbf{X}}(\hat{\mathbf{X}}^T \hat{\mathbf{X}})^{-1}\hat{\mathbf{X}}^T \\
\end{split}
\end{equation*}

In this case, 
\begin{equation*}
    \hat{\mathbf{X}}(\hat{\mathbf{X}}^T \hat{\mathbf{X}})^{-1}\hat{\mathbf{X}}^T =  V \Sigma V^T
\end{equation*}
where $V$ is an orthgonal matrix and 
\begin{equation*}
    \Sigma = \begin{bmatrix}
    1 & 0 & \dots & 0 \\
    0    & 1 & \dots & 0 \\
    \hdotsfor{4} \\
    0    &\dots & 1 & 0 \\
    \hdotsfor{4} \\
    0      & 0  & \dots & 0
\end{bmatrix}
\end{equation*}
So
\begin{equation*}
\begin{split}
\J_{\infty}(\mathbf{x}) &= \J_0(\mathbf{x})(\I_{n_0} - V \Sigma V^T) + V \Sigma V^T \\
& = \alpha \frac{1}{\sqrt{n_1}}\frac{1}{\sqrt{n_0}} \W^{(1)}\W^{(0)} (\mathbf{I}_{n_0} - V \Sigma V^T) + V \Sigma V^T \\
\end{split}
\end{equation*}
Interestingly, $(I_d - V \Sigma V^T)$ and $V \Sigma V^T$ contain orthogonal eigenvectors. For convenience,  let $\{v_i \}_{i=1}^n$ be the set of eigenvectors of $V \Sigma V^T$ with eigenvalue $1$. Furthermore, let $V_{\parallel} = \text{span}(\{v_i \}_{i=1}^n)$ and $V_{\bot} = \text{span}(\{v_i \}_{i=1}^n)^{\bot}$. Because we are in the linear region, $\mathbf{J}_{\infty}(\mathbf{x})$ and $\mathbf{J}_{0}(\mathbf{x})$ do not depend on $\mathbf{x}$. We'll use $\mathbf{J}_{\infty}$ to refer $\mathbf{J}_{\infty}(\mathbf{x})$ and $\mathbf{J}_{0}$ as $\mathbf{J}_{0}(\mathbf{x})$. 

\begin{itemize}
\item For any vector $v^{\parallel} \in V_{\parallel}$,  
\begin{equation*}
    \J_0(\I_{n_0} - V \Sigma V^T)v^{\parallel} = 0
\end{equation*}
and
\begin{equation*}
    \J_{\infty}v^{\parallel} = v^{\parallel}
\end{equation*}
Thus, all vectors in $\{v_i \}_{i=1}^n$ are eigenvetors of $\J_{\infty}$ with eigenvalue $1$ regardless of the choice of $\alpha$.

\item On the other hand, let $v$ be any complex vector such that 
\begin{equation*}
    v = \text{Re}(v) + i \text{Im}(v)
\end{equation*}
If $v$ is an eigenvector of $\J_{\infty}$ with eigenvalue $\lambda = a + ib$, then
\begin{equation*}
\begin{split}
    \J_{\infty}\text{Re}(v) &= a \text{Re}(v) - b \text{Im}(v) \\
    \J_{\infty}\text{Im}(v) &= b \text{Re}(v) + a \text{Im}(v) \\
\end{split}
\end{equation*}
Let's first decompose $\text{Re}(v)$ and $\text{Im}(v)$. 
\begin{equation*}
\begin{split}
\text{Re}(v) &= v_r^{\bot} + v_r^{\parallel} \\
\text{Im}(v) &= v_i^{\bot} + v_i^{\parallel} \\
\end{split}
\end{equation*}
where $v_r^{\bot}, v_i^{\bot} \in V_{\bot}$ and $v_r^{\parallel}, v_i^{\parallel} \in V_{\parallel}$. 
\begin{equation*}
\begin{split}
\J_{\infty}(v_r^{\bot} + v_r^{\parallel}) &= J_{0} v_r^{\bot} + v_r^{\parallel} =(a v_r^{\bot} - b v_i^{\bot}) + (a v_r^{\parallel} - b v_i^{\parallel})\\
\J_{\infty}(v_i^{\bot} + v_i^{\parallel}) &= J_{0} v_i^{\bot} + v_i^{\parallel} =(b v_r^{\bot} + a v_i^{\bot}) + (b v_r^{\parallel} + a v_i^{\parallel})\\
\end{split}
\end{equation*}
By adding and subtracting two equations,
\begin{equation*}
\begin{split}
\J_{0} (v_r^{\bot} + v_i^{\bot}) + v_r^{\parallel} + v_i^{\parallel} = \bigg[ (a + b) v_r^{\bot} + (a - b) v_i^{\bot} \bigg] + \bigg[ (a +b) v_r^{\parallel} + (a-b) v_i^{\parallel} \bigg] \\
\J_{0} (v_r^{\bot} - v_i^{\bot}) + v_r^{\parallel} - v_i^{\parallel} = \bigg[ (a - b) v_r^{\bot} - (a + b) v_i^{\bot} \bigg] + \bigg[ (a - b) v_r^{\parallel} - (a + b) v_i^{\parallel} \bigg] \\
\end{split}
\end{equation*}
When $\alpha$ is chosen such that $\Vert\J_{0}\Vert < 1$, 
\begin{equation*}
\begin{split}
    \Vert(a + b) v_r^{\bot} + (a - b) v_i^{\bot} \Vert_2 &< \Vert v_r^{\bot} + v_r^{\parallel}\Vert_2 \\
    \Vert(a - b) v_r^{\bot} - (a + b) v_i^{\bot} \Vert_2 &< \Vert v_r^{\bot} - v_r^{\parallel}\Vert_2 \\
\end{split}
\end{equation*}
Then,
\begin{equation*}
\begin{split}
    (a^2 + b^2) \Vert v_r^{\bot}\Vert_2^2 + (a^2 + b^2) \Vert v_i^{\bot}\Vert_2^2 &< \Vert v_r^{\bot}\Vert_2^2 + \Vert v_i^{\bot}\Vert_2^2 \\
    |\lambda|^2 = a^2 + b^2 < 1
\end{split}
\end{equation*}
This suggests that any complex eigenvector with components from $V_{\bot}$ would have eigenvalue with norm smaller than 1.
\end{itemize}
\end{proof}

\lemmawithbeta*
\begin{proof}
First of all, let $\mathbf{B}$ be an all-one matrix
\begin{equation*}
\begin{split}
\J_{\infty}(\mathbf{x}) &= \bigg( \hat{\mathbf{X}} - f_0(\hat{\mathbf{X}}) \bigg) \tilde{\mathbf{K}}^{-1} \frac{\partial k_x}{\partial\mathbf{x} } + \J_0(\mathbf{x})\\
& = \bigg( \hat{\mathbf{X}} - f_0(\hat{\mathbf{X}}) \bigg) \bigg(\frac{2\alpha^2}{n_0} \hat{\mathbf{X}}^T \hat{\mathbf{X}} + \beta^2\mathbf{B} \bigg)^{-1}\bigg( \frac{2\alpha^2}{n_0} \hat{\mathbf{X}}^T  \bigg) + \J_0(\mathbf{x})\\
& = \bigg( \hat{\mathbf{X}} - f_0(\hat{\mathbf{X}}) \bigg) \bigg(\hat{\mathbf{X}}^T \hat{\mathbf{X}} + \frac{n_0\beta^2}{2\alpha^2}\mathbf{B} \bigg)^{-1} \hat{\mathbf{X}}^T + \J_0(\mathbf{x})\\
&= \J_0(\mathbf{x}) + \hat{\mathbf{X}} \bigg(\hat{\mathbf{X}}^T \hat{\mathbf{X}} + \frac{n_0\beta^2}{2\alpha^2}\mathbf{B} \bigg)^{-1} \hat{\mathbf{X}}^T - ( \frac{\alpha}{\sqrt{n_1n_0}}\mathbf{W}^{(1)}\mathbf{W}^{(0)} \hat{\mathbf{X}} + \beta \frac{1}{\sqrt{n_1}} \mathbf{W}^{(1)} \mathbf{1}_{n_1} \mathbf{1}_{n}^T )  \bigg(\hat{\mathbf{X}}^T \hat{\mathbf{X}} + \frac{n_0\beta^2}{2\alpha^2}\mathbf{B} \bigg)^{-1} \hat{\mathbf{X}}^T\\
&= \J_0(\mathbf{x}) + \hat{\mathbf{X}} \bigg(\hat{\mathbf{X}}^T \hat{\mathbf{X}} + \frac{n_0\beta^2}{2\alpha^2}\mathbf{B} \bigg)^{-1} \hat{\mathbf{X}}^T - (\mathbf{J}_0(\mathbf{x})\hat{\mathbf{X}} + \beta \frac{1}{\sqrt{n_1}} \mathbf{W}^{(1)} \mathbf{1}_{n_1} \mathbf{1}_{n}^T)  \bigg(\hat{\mathbf{X}}^T \hat{\mathbf{X}} + \frac{n_0\beta^2}{2\alpha^2}\mathbf{B} \bigg)^{-1} \hat{\mathbf{X}}^T
\end{split}
\end{equation*}
Because in the linearized region, $\J_{\infty}(\mathbf{x})$ and $\J_{0}(\mathbf{x})$ do not depend on $\mathbf{x}$. We'll use $\J_{\infty}$ to refer $\J_{\infty}(\mathbf{x})$ and $\J_{0}$ as $\J_{0}(\mathbf{x})$. For simplicity, we'll also use $c = \frac{n_0\beta^2}{2\alpha^2}$. 

Based on Lemma~\ref{lemma:inverse}, 
\begin{equation*}
\begin{split}
    \hat{\mathbf{X}} \bigg(\hat{\mathbf{X}}^T \hat{\mathbf{X}} + c\mathbf{B} \bigg)^{-1} \hat{\mathbf{X}}^T = V \Lambda V^T
\end{split}
\end{equation*}
where $\Lambda = \text{diag}(\underbrace{1, ..., 1}_{n-1}, \hat{\lambda}, \underbrace{0, ..., 0}_{n_0-n})$ where $0 < \hat{\lambda} < 1$. 
Now,
\begin{equation*}
\begin{split}
    \J_{\infty} &= \J_0 ( I_{n_0} - V \Lambda V^T) + V \Lambda V^T - \beta \frac{1}{\sqrt{n_1}} W^{(1)} \mathbf{1}_{n_1} \mathbf{1}_{n}^T\bigg(\hat{\mathbf{X}}^T \hat{\mathbf{X}} + c\mathbf{B} \bigg)^{-1} \hat{\mathbf{X}}^T
\end{split}
\end{equation*}
From Corollary~\ref{cor:eigvec}, we know that the following two vectors are eigenvectors of $V\Lambda V^{T}$ with eigenvalue $\hat{\lambda}$,
\begin{equation*}
\begin{split}
    \hat{\mathbf{X}}(\hat{\mathbf{X}}^T \hat{\mathbf{X}} + c\mathbf{B} )^{-1}\mathbf{1}_{n} \qquad \hat{\mathbf{X}}(\hat{\mathbf{X}}^T \hat{\mathbf{X}} )^{-1}\mathbf{1}_{n}
\end{split}
\end{equation*}
Furthermore,
\begin{equation*}
\begin{split}
    \hat{\mathbf{X}}(\hat{\mathbf{X}}^T \hat{\mathbf{X}} + c\mathbf{B} )^{-1}\mathbf{1}_{n} = \hat{\lambda}\hat{\mathbf{X}}(\hat{\mathbf{X}}^T \hat{\mathbf{X}} )^{-1}\mathbf{1}_{n}
\end{split}
\end{equation*}
And
\begin{equation*}
    \hat{\lambda} = \frac{1}{1 + cg}
\end{equation*}
where
\begin{equation*}
    g = \text{trace}(\mathbf{B}(\hat{\mathbf{X}}^T \hat{\mathbf{X}})^{-1}) = \Vert\hat{\mathbf{X}}(\hat{\mathbf{X}}^T \hat{\mathbf{X}} )^{-1}\mathbf{1}_{n} \Vert_2^2
\end{equation*}
Let $\hat{u}$ be a rescaled unit vector of $\hat{\mathbf{X}}(\hat{\mathbf{X}}^T \hat{\mathbf{X}} )^{-1}\mathbf{1}_{n} $, then
\begin{equation*}
\begin{split}
    \J_{\infty} \hat{u} = \J_{0} (1- \hat{\lambda}) \hat{u} + \hat{\lambda} \hat{u} - \sqrt{g} \hat{\lambda} \beta \frac{1}{\sqrt{n_1}} W^{(1)} \mathbf{1}_{n_1} \hat{u}
\end{split}
\end{equation*}
\begin{equation*}
\begin{split}
    \Vert\J_{\infty} \hat{u}\Vert_2 &= \Vert \J_{0} (1- \hat{\lambda}) \hat{u} + \hat{\lambda} \hat{u} - \sqrt{g} \hat{\lambda} \beta \frac{1}{\sqrt{n_1}} \mathbf{W}^{(1)} \mathbf{1}_{n_1} \hat{u}\Vert_2 \\
    &\leq \Vert\J_{0}\Vert_{op} \Vert(1- \hat{\lambda}) \hat{u}\Vert_2 +  \Vert \hat{\lambda} \hat{u} \Vert_2 + \Vert \sqrt{g} \hat{\lambda}  \beta \frac{1}{\sqrt{n_1}} \mathbf{W}^{(1)} \mathbf{1}_{n_1} \hat{u}\Vert_2 \\
    & = (1- \hat{\lambda}) \Vert\J_{0}\Vert_{op} + \hat{\lambda} +  \sqrt{g} \hat{\lambda} \Vert  \beta \frac{1}{\sqrt{n_1}} \mathbf{W}^{(1)} \mathbf{1}_{n_1}\Vert_2 \\
    & < (1- \hat{\lambda}) (1 - \Delta) + \hat{\lambda} +  \sqrt{g} \hat{\lambda}\frac{\beta^2 n_0 \Delta}{2r\alpha^2}\\
    & \leq (1- \hat{\lambda}) (1 - \Delta) + \hat{\lambda} +  g \hat{\lambda}\frac{\beta^2 n_0 \Delta}{2\alpha^2} \qquad (Lemma~\ref{lemma:lower_bound})\\
    & = \frac{(1 - \Delta)cg + 1 + \frac{g\beta^2n_0\Delta}{2\alpha^2}}{1 + cg} = 1
\end{split}
\end{equation*}
Therefore, $\J_{\infty}$ will shrink every vectors orthogonal to the eigenvectors in $V$ with eigenvalue $1$. By the same arguments in the proof of Lemma~\ref{lemma:no-beta}, we can conclude the proof. 
\end{proof}

\subsection{Useful Lemmas}
\begin{lemmma}
\label{lemma:inverse}
Suppose $\mathbf{X} \in \R^{k \times m}$ is a full-rank matrix with $k \geq m$ and $m \geq 2$. Let $c$ be an arbitrary positive constant and $\mathbf{B}$ an all-one matrix. Consider the following real symmetric matrix, 
\begin{equation*}
    \mathbf{X} (\mathbf{X}^T \mathbf{X} + c \mathbf{B})^{-1} \mathbf{X}^T
\end{equation*}
It can be characterized by having eigenvalue $1$ with multiplicity $m-1$, eigenvalue $0$ with multiplicity $k-m$ and another eigenvalue $\lambda$ such that $0 < \lambda < 1$. 

\end{lemmma}

\begin{proof}
By \citep{miller1981inverse}, if $P$ and $P+Q$ are invertible, and $Q$ has rank 1, then let $g' = \text{trace}(QP^{-1})$, we know that $g' \neq 1$, and
\begin{equation*}
    (P + Q)^{-1} = P^{-1} - \frac{1}{1 + g'}P^{-1} Q P^{-1}
\end{equation*}
First of all, it is easy to see that $(\mathbf{X}^T \mathbf{X} + c \mathbf{B})^{-1}$ is invertible. This is because $\mathbf{X}^T \mathbf{X}$ is positive definite and $cB$ is positive semi-definite.  

Since $B$ is a rank one matrix, 
\begin{equation*}
    (\mathbf{X}^T \mathbf{X} + c \mathbf{B})^{-1} = \underbrace{(\mathbf{X}^T \mathbf{X})^{-1}}_{I_1} - \underbrace{\frac{c}{1 + cg}(\mathbf{X}^T \mathbf{X})^{-1} \mathbf{B} (\mathbf{X}^T \mathbf{X})^{-1}}_{I_2}
\end{equation*}
where $g = \text{trace}(\mathbf{B}(\mathbf{X}^T \mathbf{X})^{-1})$.

Let's consider the singular value decomposition of $\mathbf{X}^T = U \Sigma V^T$ 

\begin{itemize}
    \item 

$\mathbf{X}^T \mathbf{X} = U \Sigma^2 U^T$ and $(\mathbf{X}^T \mathbf{X})^{-1} = U \Sigma^{-2} U^T$. So
\begin{equation*}
    \mathbf{X} I_1 \mathbf{X}^T = \mathbf{X} (\mathbf{X}^T \mathbf{X})^{-1} \mathbf{X}^T = V \Sigma U^T U \Sigma^{-2} U^T U \Sigma V^T = V \Lambda_m V^T
\end{equation*}
where $\Lambda_m = \text{diag}(\underbrace{1, ..., 1}_{m}, \underbrace{0, ..., 0}_{k-m})$

    \item 
    \begin{equation*}
    \begin{split}
        \mathbf{X}^T I_2 \mathbf{X} = \frac{c}{1 + cg}M = \frac{c}{1 + cg} \mathbf{X} (\mathbf{X}^T \mathbf{X})^{-1} \mathbf{B} (\mathbf{X}^T \mathbf{X})^{-1}\mathbf{X}^T
    \end{split}
    \end{equation*}
    The first thing to notice is that $\mathbf{B} = \textbf{1}  \textbf{1}^T$ where $ \textbf{1}$ is a vector of ones. Therefore,
    \begin{equation*}
        M = \mathbf{X} (\mathbf{X}^T \mathbf{X})^{-1} \mathbf{B} (\mathbf{X}^T \mathbf{X})^{-1} \mathbf{X}^T = \mathbf{X}(\mathbf{X}^T \mathbf{X})^{-1}  \textbf{1}  \textbf{1}^T (\mathbf{X}^T \mathbf{X})^{-1} \mathbf{X}^T = \mathbf{a}\mathbf{a}^T
    \end{equation*}
    where $\mathbf{a} = \mathbf{X} (\mathbf{X}^T \mathbf{X})^{-1}\textbf{1}$.
    
    This implies that $M$ is a rank one matrix with singular value $\Vert\mathbf{a}\Vert^2$. But we also know the following:
    \begin{equation*}
        \begin{split}
            \Vert\mathbf{a}\Vert^2 = \mathbf{a}^T  \mathbf{a} &= \text{trace} (\mathbf{a} \mathbf{a}^T) = \text{trace} (M) \\
            & = \text{trace} (\mathbf{X} (\mathbf{X}^T \mathbf{X})^{-1} \mathbf{B} (\mathbf{X}^T \mathbf{X})^{-1} \mathbf{X}^T) = \text{trace} (\mathbf{X}^T \mathbf{X} (\mathbf{X}^T \mathbf{X})^{-1} \mathbf{B} (\mathbf{X}^T \mathbf{X})^{-1}) \\
            & = \text{trace} (\mathbf{B} (\mathbf{X}^T \mathbf{X})^{-1}) = g > 0
        \end{split}
    \end{equation*}
    The last strict inequality comes from the fact that $X$ is full rank so that $\mathbf{X} (\mathbf{X}^T \mathbf{X})^{-1}$ has no zero singular value. Furthermore, 
    \begin{equation*}
    \begin{split}
        \mathbf{X} I_1 \mathbf{X}^T \mathbf{a} &= \mathbf{X}(\mathbf{X}^T \mathbf{X})^{-1}\mathbf{X}^T \mathbf{a} \\
        &= \mathbf{X}(\mathbf{X}^T \mathbf{X})^{-1}\mathbf{X}^T \mathbf{X} (\mathbf{X}^T \mathbf{X})^{-1}\textbf{1} = \mathbf{X} (\mathbf{X}^T \mathbf{X})^{-1}\textbf{1} \\
        &= \mathbf{a}
    \end{split}
    \end{equation*}
    Because $\mathbf{a}$ is not a zero vector, it is also one of the eigenvector of $\mathbf{X} I_1 X$ with eigenvalue $1$. 
    
    And the eigenvalue of $X^T I_2 X$ is the following:
    \begin{equation*}
        \begin{split}
            0 < \frac{cg}{1 + cg} < 1
        \end{split}
    \end{equation*}
    The inequalities comes from the fact that $c$ is also non-negative. We'll denote $\sigma = \frac{cg}{1 + cg}$. So
    \begin{equation*}
        \mathbf{X} I_2 \mathbf{X}^T = \sigma \hat{\mathbf{a}}\hat{\mathbf{a}}^T
    \end{equation*}
    where $\hat{\mathbf{a}}$ is $\mathbf{a}$ rescaled to have unit length. 
    
\end{itemize}

Now that we have examined two parts separately. Let's put them together. For convenience, we'll also denote $\mathbf{X} (\mathbf{X}^T \mathbf{X} + c \mathbf{B})^{-1} \mathbf{X}^T = \mathbf{X} I_1 \mathbf{X}^T - \mathbf{X} I_2 \mathbf{X}^T = \mathbf{M}_1 - \mathbf{M}_2$.

Based on the eigen decomposition of $\mathbf{M}_1$,
\begin{equation*}
\begin{split}
    \mathbf{M}_1 = \sum_{k=1}^m \mathbf{u}_k \mathbf{u}_k^T
\end{split}
\end{equation*}
with lost of generality, let's also denote $\hat{\mathbf{a}} = \mathbf{u}_1$. Now,
\begin{equation*}
\begin{split}
    \mathbf{M}_1 - \mathbf{M}_2 &= \sum_{k=1}^m \mathbf{u}_k \mathbf{u}_k^T - \sigma \mathbf{u}_1 \mathbf{u}_1^T \\
    &= (1 - \sigma) \mathbf{u}_1 \mathbf{u}_1^T + \sum_{k=2}^m \mathbf{u}_k \mathbf{u}_k^T
\end{split}
\end{equation*}
Because $0 < \sigma < 1$, $\mathbf{X} (\mathbf{X}^T \mathbf{X} + c \mathbf{B})^{-1} \mathbf{X}^T$ has eigenvalue $1$ with multiplicity $m-1$, eigenvalue $0$ with multiplicity $k-m$ and another eigenvalue $\lambda$ such that $0 < \lambda < 1$. 
\end{proof}

\begin{corollary}
\label{cor:eigvec}
Following the setup in Lemma~\ref{lemma:inverse}, we could also know that $\mathbf{X}^T (\mathbf{X}^T \mathbf{X})^{-1}\mathbf{1}$ is an eigenvector with with eigenvalue $\lambda$ and 
\begin{equation*}
    \bigg ( \mathbf{X} (\mathbf{X}^T \mathbf{X} + c \mathbf{B})^{-1} \mathbf{X}^T \bigg) \mathbf{X} (\mathbf{X}^T \mathbf{X})^{-1}\mathbf{1} =  \mathbf{X} (\mathbf{X}^T \mathbf{X} + c \mathbf{B})^{-1} \mathbf{1} = \lambda \mathbf{X} (\mathbf{X}^T \mathbf{X})^{-1}\mathbf{1}
\end{equation*}
\end{corollary}

\begin{corollary}
\label{cor:inverse_op}
Suppose $\mathbf{X} \in \R^{k \times m}$ is a full-rank matrix with $k \geq m$ and $m \geq 2$. Let $c$ be an arbitrary non-negative constant and $\mathbf{B}$ an all-one matrix. 
\begin{equation*}
    \Vert\mathbf{X} (\mathbf{X}^T \mathbf{X} + c \mathbf{B})^{-1} \mathbf{X}^T\Vert_{op} = 1
\end{equation*}
\end{corollary}

\begin{remark}
$c$ can also takes on negative values as long as $cg$ is not close to $-1$.
\end{remark}

\begin{lemmma}
\label{lemma:lower_bound}
Suppose $\mathbf{X} \in \R^{k \times m}$ is a full-rank matrix with $k \geq m$ and $\mathbf{B}$ an all-one matrix. If 
\begin{equation*}
    \Vert\mathbf{X}_{\cdot, i}\Vert_2 = r \qquad \forall i \in [m]
\end{equation*}
Then,
\begin{equation*}
    \mathrm{trace}(\mathbf{B}(\mathbf{X}^T\mathbf{X})^{-1}) \geq \frac{1}{r^2}
\end{equation*}
\end{lemmma}
\begin{proof}
First of all,
\begin{equation*}
\begin{split}
     \mathrm{trace}(\mathbf{B}(\mathbf{X}^T\mathbf{X})^{-1}) &\geq \mathrm{trace}(\mathbf{1}^T(\mathbf{X}^T\mathbf{X})^{-1}\mathbf{1}) \\
     & \geq \Vert\mathbf{1}\Vert_2^2 \frac{1}{\Vert\mathbf{X}^T\mathbf{X}\Vert_{op}} = \frac{m}{\Vert\mathbf{X}^T\mathbf{X}\Vert_{op}} 
\end{split}
\end{equation*}
On the hand,
\begin{equation*}
\begin{split}
    \Vert\mathbf{X}^T\mathbf{X}\Vert_{op} = \Vert\mathbf{X}^T\Vert_{op}^2 \leq \Vert\mathbf{X}^T\Vert_{f}^2 \leq \mathrm{trace}(\mathbf{X}^T\mathbf{X}) \leq r^2m
\end{split}
\end{equation*}
Therefore,
\begin{equation*}
    \mathrm{trace}(\mathbf{B}(\mathbf{X}^T\mathbf{X})^{-1}) \geq \frac{1}{r^2}
\end{equation*}
\end{proof}

\section{Proofs for Sec~\ref{sec:mutil-beyond-linear}}
\subsection{Derivation for the Approximated NTK}
\label{append:deri-ap-ntk}
The closed form NTK of $\mathrm{erf}$ \cite{lee2019wide, williams1997computing} can be written with the following two components:
\begin{equation*}
    \begin{split}
        \mathcal{T}(\Sigma, \text{erf}, \text{erf})(\mathbf{x}, \hat{\mathbf{x}}) &= \frac{2}{\pi} \arcsin\bigg( \frac{\Sigma(\mathbf{x}, \hat{\mathbf{x}})}{\sqrt{( \Sigma(\mathbf{x}, \mathbf{x}) + 0.5 )(\Sigma(\hat{\mathbf{x}}, \hat{\mathbf{x}}) + 0.5)}}\bigg) \\
        \mathcal{T}(\Sigma, \dot{\text{erf}}, \dot{\text{erf}})(\mathbf{x}, \hat{\mathbf{x}}) &= \frac{4}{\pi}\det(I + 2\Sigma )^{-\frac{1}{2}} = \frac{4}{\pi} \frac{1}{\sqrt{(1 +2\Sigma(\mathbf{x}, \mathbf{x})(1 +2\Sigma(\hat{\mathbf{x}}, \hat{\mathbf{x}})) - 4\Sigma(\mathbf{x}, \hat{\mathbf{x}})^2 ) }}
    \end{split}
\end{equation*}

Here, we can approximate $\mathrm{sigmoid}$ function $\sigma_s$ by $\mathrm{erf}$ function: 
\begin{equation*}
    \sigma_s(x)  \approx  \sigma_{\hat{s}}(x) = \frac{1}{2}\text{erf}(\frac{1}{2}x) + \frac{1}{2}
\end{equation*}
Then, 
\begin{equation*}
\begin{split}
\mathcal{T}(\Sigma, \sigma_{\hat{s}}, \sigma_{\hat{s}})(\mathbf{x}, \hat{\mathbf{x}}) = \Ex_{u, v \sim \mathcal{N}(0, \Sigma)} [\sigma_{\hat{s}}(u)\sigma_{\hat{s}}(v)] &= \Ex[\frac{1}{4}\text{erf}(\frac{1}{2}u)\text{erf}(\frac{1}{2}v)] + \Ex[\frac{1}{4}\text{erf}(\frac{1}{2}u) + \frac{1}{4}\text{erf}(\frac{1}{2}v)] + \frac{1}{4}\\
& = \frac{1}{4}\Ex[\text{erf}(\frac{1}{2}u)\text{erf}(\frac{1}{2}v)]  + \frac{1}{4} \\
&= \frac{1}{4}\mathcal{T}(\frac{1}{4}\Sigma, \text{erf}, \text{erf})(\mathbf{x}, \hat{\mathbf{x}}) + \frac{1}{4} 
\end{split}
\end{equation*}
\begin{equation*}
\begin{split}
\boxed{\mathcal{T}(\Sigma, \sigma_{\hat{s}}, \sigma_{\hat{s}})(\mathbf{x}, \hat{\mathbf{x}}) = \frac{1}{4} + \frac{1}{2\pi}\arcsin\bigg( \frac{\Sigma(\mathbf{x}, \hat{\mathbf{x}})}{\sqrt{( \Sigma(\mathbf{x}, \mathbf{x}) + 2)(\Sigma(\hat{\mathbf{x}}, \hat{\mathbf{x}}) + 2)}}\bigg)}
\end{split}
\end{equation*}
and 
\begin{equation*}
\begin{split}
\mathcal{T}(\Sigma, \dot{\sigma}_{\hat{s}}, \dot{\sigma}_{\hat{s}})(\mathbf{x}, \hat{\mathbf{x}}) &= \Ex_{u, v \sim \mathcal{N}(0, \Sigma)} [\dot{\sigma}_{\hat{s}}(u)\dot{\sigma}_{\hat{s}}(v)] = \frac{1}{16}\Ex[\dot{\text{erf}}(\frac{1}{2}u)\dot{\text{erf}}(\frac{1}{2}v)] \\
& =\frac{1}{16} \mathcal{T}(\frac{1}{4}\Sigma, \dot{\text{erf}}, \dot{\text{erf}})(\mathbf{x}, \hat{\mathbf{x}})
\end{split}
\end{equation*}
\begin{equation*}
\boxed{\mathcal{T}(\Sigma, \dot{\sigma}_{\hat{s}}, \dot{\sigma}_{\hat{s}})(\mathbf{x}, \hat{\mathbf{x}}) = \frac{1}{2\pi} \frac{1}{\sqrt{(2 +\Sigma(\mathbf{x}, \mathbf{x})(2 +\Sigma(\hat{\mathbf{x}}, \hat{\mathbf{x}})) - \Sigma(\mathbf{x}, \hat{\mathbf{x}})^2 ) }}}
\end{equation*}

Based on the definition of NTK, we can derive the following for ${\sigma}_{\hat{s}}$
\begin{equation*}
\begin{split}
\Theta^{1}_{\infty}(\hat{\mathbf{x}}, \mathbf{x}) &= \Sigma^{1}(\hat{\mathbf{x}}, \mathbf{x}) =  \frac{1}{n_0} \hat{\mathbf{x}}^T \mathbf{x} \\
\Theta^{2}_{\infty}(\hat{\mathbf{x}}, \mathbf{x}) &=  \Theta^{1}_{\infty}(\hat{\mathbf{x}}, \mathbf{x}) \mathcal{T}(\Theta^{1}_{\infty}, \dot{\sigma}_{\hat{s}}, \dot{\sigma}_{\hat{s}})(\mathbf{x}, \hat{\mathbf{x}}) + \mathcal{T}(\Theta^{1}_{\infty}, \sigma_{\hat{s}}, \sigma_{\hat{s}})(\mathbf{x}, \hat{\mathbf{x}})
\end{split}
\end{equation*}
Let's look at the first part
\begin{equation*}
\begin{split}
\Theta^{1}_{\infty}(\hat{\mathbf{x}}, \mathbf{x}) \mathcal{T}(\Theta^{1}_{\infty}, \dot{\sigma}_{\hat{s}}, \dot{\sigma}_{\hat{s}})(\mathbf{x}, \hat{\mathbf{x}}) &= \frac{1}{2\pi} \frac{1}{\sqrt{(2 +\frac{1}{n_0} \mathbf{x}^T \mathbf{x})(2 +\frac{1}{n_0} \hat{\mathbf{x}}^T \hat{\mathbf{x}} ) - (\frac{1}{n_0} \hat{\mathbf{x}}^T \mathbf{x})^2 )}}[\frac{1}{n_0} \hat{\mathbf{x}}^T \mathbf{x}] \\
& = \frac{1}{2\pi} \frac{\hat{\mathbf{x}}^T \mathbf{x}}{\sqrt{(2n_0 + \mathbf{x}^T \mathbf{x})(2n_0 + \hat{\mathbf{x}}^T \hat{\mathbf{x}} ) - (\hat{\mathbf{x}}^T \mathbf{x})^2 )}} 
\end{split}
\end{equation*}
and the second part
\begin{equation*}
\begin{split}
\mathcal{T}(\Theta^{1}_{\infty}, \sigma_{\hat{s}}, \sigma_{\hat{s}})(\mathbf{x}, \hat{\mathbf{x}}) &= \frac{1}{4} + \frac{1}{2\pi}\arcsin\bigg( \frac{\frac{1}{n_0} \hat{\mathbf{x}}^T \mathbf{x}}{\sqrt{( \frac{1}{n_0} {\mathbf{x}}^T \mathbf{x} + 2)(\frac{1}{n_0} \hat{\mathbf{x}}^T \hat{\mathbf{x}}+ 2)}}\bigg)\\
& = \frac{1}{4} + \frac{1}{2\pi}\arcsin\bigg( \frac{\hat{\mathbf{x}}^T \mathbf{x}}{\sqrt{( {\mathbf{x}}^T \mathbf{x} + 2n_0)( \hat{\mathbf{x}}^T \hat{\mathbf{x}}+ 2n_0)}}\bigg)\\
\end{split}
\end{equation*}

\subsection{Detailed Discussion of $\frac{\partial \mathbf{k}_x}{\partial\mathbf{x} }$}\label{app_dkdx}
Without loss of generality, we will focus on $\frac{\partial \mathbf{k}_x}{\partial\mathbf{x} }\rvert_{\mathbf{x}_1}$,
\label{sec:grad-component}
$$\frac{\partial \mathbf{k}_x}{\partial\mathbf{x} } = 
\begin{bmatrix}
\frac{\Theta_{\infty}^{L}(\mathbf{x}_1, \mathbf{x})}{\partial\mathbf{x}} \\
\dots\\
\frac{\Theta_{\infty}^{L}(\mathbf{x}_n, \mathbf{x})}{\partial\mathbf{x}}
\end{bmatrix}$$ where
\begin{equation*}
    \frac{\partial \Theta_{\infty}^{L}(\hat{\mathbf{x}}, \mathbf{x})}{\partial\mathbf{x}}  = \underbrace{\frac{\partial \mathcal{T}(\Theta^{1}_{\infty}, \sigma_{\hat{s}}, \sigma_{\hat{s}})(\hat{\mathbf{x}}, \mathbf{x}))}{\partial\mathbf{x}}}_{I^g_1(\hat{\mathbf{x}}, \mathbf{x})}  + \underbrace{\frac{\partial\Theta^{1}_{\infty}(\hat{\mathbf{x}}, \mathbf{x})) \mathcal{T}(\Theta^{1}_{\infty}, \dot{\sigma}_{\hat{s}}, \dot{\sigma}_{\hat{s}})(\hat{\mathbf{x}}, \mathbf{x})) }{\partial\mathbf{x}}}_{I^g_2(\hat{\mathbf{x}}, \mathbf{x})} 
\end{equation*}

\begin{figure}[t]
\includegraphics[width=0.25\linewidth]{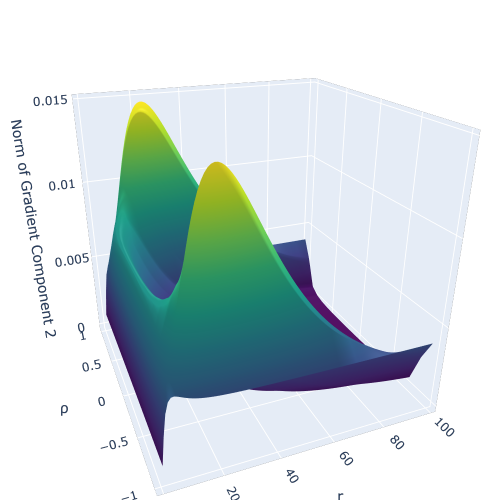}
\centering
\caption{$\rho$,  $r$ vs Norm of Gradient Component 2}
\label{fig:grdient-component}
\end{figure}

Let's look at each row separately, and break this down into two parts.
\begin{itemize}
    \item $I^g_1(\hat{\mathbf{x}}, \mathbf{x})$ 
    
    After deriving the derivative, we get this:
    \begin{equation*}
    \begin{split}
        I^g_1(\hat{\mathbf{x}}, \mathbf{x}) &= \frac{1}{2\pi} \frac{1}{\sqrt{1 - A^2}} \frac{\hat{\mathbf{x}}\bigg[(\hat{\mathbf{x}}^T\hat{\mathbf{x}}+2 n_0)({\mathbf{x}}^T{\mathbf{x}}+2 n_0) \bigg] - \mathbf{x} \bigg[(\hat{\mathbf{x}}^T\hat{\mathbf{x}}+2 n_0){\mathbf{x}}^T\hat{\mathbf{x}}\bigg] }{\bigg[ (\hat{\mathbf{x}}^T\hat{\mathbf{x}}+2 n_0)({\mathbf{x}}^T{\mathbf{x}}+2 n_0)\bigg]^{\frac{3}{2}}} \\
        A &= \frac{\hat{\mathbf{x}}^T \mathbf{x}}{\sqrt{( {\mathbf{x}}^T \mathbf{x} + 2n_0)( \hat{\mathbf{x}}^T \hat{\mathbf{x}}+ 2n_0)}}
    \end{split}
    \end{equation*}
    Since we are only interested in $\mathbf{J}_{\infty}(\mathbf{x}_1)$ and each row of $\frac{\partial \mathbf{k}_x}{\partial\mathbf{x} }$, we'll examine $I^g_1(\mathbf{x}_i, \mathbf{x}_1)$. 
    \begin{equation*}
    \begin{split}
        I^g_1(\mathbf{x}_i, \mathbf{x}_1) &= \frac{1}{2\pi} \frac{r^2 + 2n_0}{\sqrt{(r^2+2n_0)^2 - (r^2\rho_{i1})^2}} \frac{\mathbf{x}_i(r^2 + 2n_0)^2 - \mathbf{x}_1 \bigg[(r^2 + 2n_0)r^2\rho_{i1} \bigg]}{(r^2+2n_0)^3} \\
        & = \frac{1}{2\pi} \frac{1}{\sqrt{(r^2+2n_0)^2 - (r^2\rho_{i1})^2}} \frac{\mathbf{x}_i(r^2 + 2n_0) - \mathbf{x}_1 r^2\rho_{i1}}{r^2+2n_0}
    \end{split}
    \end{equation*}
    It is easy to see that $I^g_1(\mathbf{x}_i, \mathbf{x}_1) \to 0$ as $r$ grows regardless of $\rho_{i1}$.

    \item $I^g_2(\hat{\mathbf{x}}, \mathbf{x})$
    
    We know that
    \begin{equation*}
    \begin{split}
        I^g_2(\hat{\mathbf{x}}, \mathbf{x}) &= \frac{1}{2\pi}\frac{\hat{\mathbf{x}} \bigg[ ( {\mathbf{x}}^T \mathbf{x} + 2n_0)( \hat{\mathbf{x}}^T \hat{\mathbf{x}}+ 2n_0) - (\hat{\mathbf{x}}^T \mathbf{x})^2\bigg] - \hat{\mathbf{x}}^T \mathbf{x} \bigg[ (2n_0 + \hat{\mathbf{x}}^T \hat{\mathbf{x}})\mathbf{x} -  (\hat{\mathbf{x}}^T \mathbf{x})\hat{\mathbf{x}}  \bigg] }{\bigg[( {\mathbf{x}}^T \mathbf{x} + 2n_0)( \hat{\mathbf{x}}^T \hat{\mathbf{x}}+ 2n_0) - (\hat{\mathbf{x}}^T \mathbf{x})^2 \bigg]^{\frac{3}{2}}}
    \end{split}
    \end{equation*}
    Again, let's examine $I^g_2(\mathbf{x}_i, \mathbf{x}_1)$. 
    \begin{equation*}
    \begin{split}
    I^g_2(\mathbf{x}_i, \mathbf{x}_1) &= \frac{1}{2\pi} \frac{(r^2 + 2n_0)^2\mathbf{x}_i  - r^2\rho_{i1}(2n_0 + r^2)\mathbf{x}_1 }{\bigg[(r^2 + 2n_0)^2 - r^4\rho_{i1}^2 \bigg]^{\frac{3}{2}}}
    \end{split}
    \end{equation*}
\begin{equation*}
\begin{split}
\Vert I^g_2(\mathbf{x}_i, \mathbf{x}_1)\Vert_2^2 &= \frac{1}{4\pi^2} \frac{r^2\bigg[(r^2 + 2n_0)^4 + r^4\rho_{i1}^2(2n_0 + r^2)^2 - 2 r^2\rho_{i1}^2(2n_0 + r^2)^3\bigg]}{\bigg[(r^2 + 2n_0)^2 - r^4\rho_{i1}^2 \bigg]^{3}} \\
& = \frac{1}{4\pi^2} \frac{16n_0^4 + r^2\Bigg[n_0^3(32 - 16\rho_{i1}^2) + r^2\bigg[ n_0^2(24 - 20\rho_{i1}^2) + r^2\big[n_0(8 - 8\rho_{i1}^2) + r^2(1 - \rho_{i1}^2)\big]\bigg]\Bigg]}{\bigg[r^4(1-\rho_{i1}^2) + 4n_0r^2 + 4n_0^2 \bigg]^3}
\end{split}
\end{equation*}
Based on the equation for $\Vert I^g_2(\mathbf{x}_i, \mathbf{x}_1)\Vert_2^2$, we know that if $\rho_{i1}^2 \neq 1$, $\Vert I^g_2(\mathbf{x}_i, \mathbf{x}_1)\Vert_2^2$ eventually decays to zero  with larger $r$. But $\Vert I^g_2(\mathbf{x}_i, \mathbf{x}_1)\Vert_2^2$ converges to a constant if $\rho_{i1}^2 = 1 $. For simplicity, in this section, we do not assume there is any parallel input. Therefore, we can see that all the other terms will go to zero except $I^g_2(\mathbf{x}_1, \mathbf{x}_1)$. It is worth noting that if $\rho_{i1}$ is close to one, the norm will see a spike before going down to zero. But in practice, the data is more than likely to be well separated with small $|\rho_{ij}|$. The discussion here is illustrated in Figure.~\ref{fig:grdient-component}. 
\end{itemize}

Combining the above analysis on the two components of gradient, it is easy to see that with large $r$, 
\begin{equation*}
\begin{split}
\frac{\partial \mathbf{k}_x}{\partial\mathbf{x} }\rvert_{\mathbf{x}_1} \approx
\begin{bmatrix}
\frac{\Theta_{\infty}^{L}(\mathbf{x}_1, \mathbf{x})}{\partial\mathbf{x}}\rvert_{\mathbf{x}_1} \\
\mathbf{0} \\
\dots\\
\mathbf{0} 
\end{bmatrix} 
\end{split}
\end{equation*}
\begin{equation*}
\begin{split}
    \Vert \frac{\partial \mathbf{k}_x}{\partial\mathbf{x} }\rvert_{\mathbf{x}_1} \Vert_{op} \approx \Vert\frac{\Theta_{\infty}^{L}(\mathbf{x}_1, \mathbf{x})}{\partial\mathbf{x}}\rvert_{\mathbf{x}_1} \Vert_2 \approx \Vert \frac{1}{2\pi}\frac{2n_0(r^2 + 2n_0)}{(4n_0r^2 + 4n_0^2)^{\frac{3}{2}}} \mathbf{x}_1 \Vert_2 = \frac{1}{2\pi} \frac{2n_0r(r^2 + 2n_0)}{(4n_0r^2 + 4n_0^2)^{\frac{3}{2}}} \approx \frac{1}{8\pi \sqrt{n_0}}
\end{split}
\end{equation*}

\subsection{Parallel Inputs Analysis}
\label{appnd:parrel-inputs}
In the previous section, we assume that there are no parallel inputs. But this assumption is not necessary. In fact,  given training data $\{ \mathbf{x}_i \}_1^n$, w.l.o.g, let's impose $ \mathbf{x}_1 = -  \mathbf{x}_2$. Based on the results we have in Section~\ref{sec:mutil-beyond-linear}, we can still derive a similar approximation for the NTK regression solution.
\begin{itemize}
\item $\tilde{\mathbf{K}}$ 

Fisrt of all,
\begin{equation*}
\begin{split}
\mathbf{K}^1_{ij} = \mathcal{T}(\Theta^{1}_{\infty}, \sigma_{\hat{s}}, \sigma_{\hat{s}})(\mathbf{x}_i, \mathbf{x}_j) 
& = \frac{1}{4} + \frac{1}{2\pi}\arcsin\bigg( \frac{{r^2\rho_{i,j}}}{(r^2 + 2n_0)}\bigg)\\
\end{split}
\end{equation*}
If $\rho_{i,j} = 1$, then $\mathbf{K}^1_{ij}$ is going to converge to $\frac{1}{2}$ as $r$ grows bigger. But if $\rho_{i,j} = -1$, this term is going to zero. 

Therefore, $\tilde{\mathbf{K}}$ can be approximated by this block diagonal matrix. 
\begin{equation*}
\begin{split}
\tilde{\mathbf{K}} \approx
\begin{bmatrix}
B_1 &  \dots &  0\\
0 & B_2  & 0 \\
\dots & \dots & \dots\\
0 & \dots & B_2
\end{bmatrix}
\end{split}
\end{equation*}
where 
\begin{equation*}
\begin{split}
B_1 = 
\begin{bmatrix}
I_k + \frac{1}{2} & -I_k \\
-I_k & I_k + \frac{1}{2}
\end{bmatrix}
\qquad
B_2 = I_k + \frac{1}{2}
\qquad
I_k = \frac{1}{2\pi} \frac{r^2}{\sqrt{4n_0^2 + 4n_0r^2}} \approx \frac{r}{4\pi \sqrt{n_0}}
\end{split}
\end{equation*}
The inverse of $\tilde{\mathbf{K}} $, is the following, as r grows large:
\begin{equation*}
\begin{split}
\tilde{\mathbf{K}}^{-1} \approx
\begin{bmatrix}
B_1^{-1} &  \dots &  0\\
0 & \frac{1}{I_k + \frac{1}{2}}  & 0 \\
\dots & \dots & \dots\\
0 & \dots & \frac{1}{I_k + \frac{1}{2}} 
\end{bmatrix}
\approx 
\begin{bmatrix}
B_1^{-1} &  \dots &  0\\
0 & \frac{4\pi \sqrt{n_0}}{r}  & 0 \\
\dots & \dots & \dots\\
0 & \dots & \frac{4\pi \sqrt{n_0}}{r}
\end{bmatrix}
\end{split}
\end{equation*}
where 
\begin{equation*}
    B_1^{-1} = \frac{1}{I_k + \frac{1}{4}} 
\begin{bmatrix}
I_k + \frac{1}{2} & I_k \\
I_k & I_k + \frac{1}{2}
\end{bmatrix}
\end{equation*} 
\item $\frac{\partial \mathbf{k}_x}{\partial\mathbf{x} }$

Based on the discussion from Section~\ref{sec:mutil-beyond-linear}, 
\begin{equation*}
\frac{\partial \mathbf{k}_x}{\partial\mathbf{x} }\rvert_{\mathbf{x}_1} \approx
\begin{bmatrix}
\frac{\Theta_{\infty}^{L}(\mathbf{x}_1, \mathbf{x})}{\partial\mathbf{x}}\rvert_{\mathbf{x}_1} \\
-\frac{\Theta_{\infty}^{L}(\mathbf{x}_1, \mathbf{x})}{\partial\mathbf{x}}\rvert_{\mathbf{x}_1}  \\
\dots\\
\mathbf{0} 
\end{bmatrix}
= \begin{bmatrix}
J_k \mathbf{x}_1\\
-J_k \mathbf{x}_1 \\
\dots\\
\mathbf{0} 
\end{bmatrix}
\end{equation*}
where
\begin{equation*}
    J_k = \frac{1}{2\pi}\frac{2n_0(r^2 + 2n_0)}{(4n_0r^2 + 4n_0^2)^{\frac{3}{2}}} \approx \frac{1}{8\pi \sqrt{n_0}} \frac{1}{r} 
\end{equation*}
\end{itemize}

Finally, 
\begin{equation*}
\begin{split}
\bigg( \hat{\mathbf{X}} - f_0(\hat{\mathbf{X}}) \bigg) \tilde{\mathbf{K}}^{-1} \frac{\partial k_x}{\partial\mathbf{x} }  &\approx \hat{\mathbf{X}}\tilde{\mathbf{K}}^{-1} \frac{\partial k_x}{\partial\mathbf{x} } \approx \hat{\mathbf{X}} \begin{bmatrix}
B_1^{-1} &  \dots &  0\\
0 & \frac{1}{I_k + \frac{1}{2}}  & 0 \\
\dots & \dots & \dots\\
0 & \dots & \frac{1}{I_k + \frac{1}{2}}
\end{bmatrix}
\begin{bmatrix}
J_k \mathbf{x}_1\\
-J_k \mathbf{x}_1 \\
\mathbf{0} \\
\dots\\
\mathbf{0} 
\end{bmatrix} \\
& = \hat{\mathbf{X}} 
\begin{bmatrix}
\frac{\frac{1}{2} J_k}{I_k + \frac{1}{4}} \mathbf{x}_1\\
-\frac{\frac{1}{2} J_k}{I_k + \frac{1}{4}}  \mathbf{x}_1 \\
\mathbf{0} \\
\dots\\
\mathbf{0} 
\end{bmatrix}  = 2\frac{\frac{1}{2} J_k}{I_k + \frac{1}{4}}\mathbf{x}_1\mathbf{x}_1^T
\end{split}
\end{equation*}
Thus,
\begin{equation*}
\Vert \bigg( \hat{\mathbf{X}} - f_0(\hat{\mathbf{X}}) \bigg) \tilde{\mathbf{K}}^{-1} \frac{\partial \mathbf{k}_x}{\partial\mathbf{x} } \Vert_{op} 
\approx \frac{r^2J_k}{I_k + \frac{1}{4}} 
= \frac{r^2 \frac{1}{8\pi \sqrt{n_0}} \frac{1}{r} }{ \frac{r}{4 \pi \sqrt{n_0}}} = \frac{1}{2}
\end{equation*}
By similar argument, as $r \to \infty$, we have
\begin{equation*}
    \Vert\J_{\infty}(\mathbf{x})\Vert_{op} \leq \frac{1}{2}
\end{equation*}

\clearpage
\section{Additional Simulations}
\subsection{Multiple Points: Linear Region}
\begin{figure*}
    \centering
    \begin{subfigure}{0.3\linewidth}
        \centering
        \includegraphics[height=1.2in]{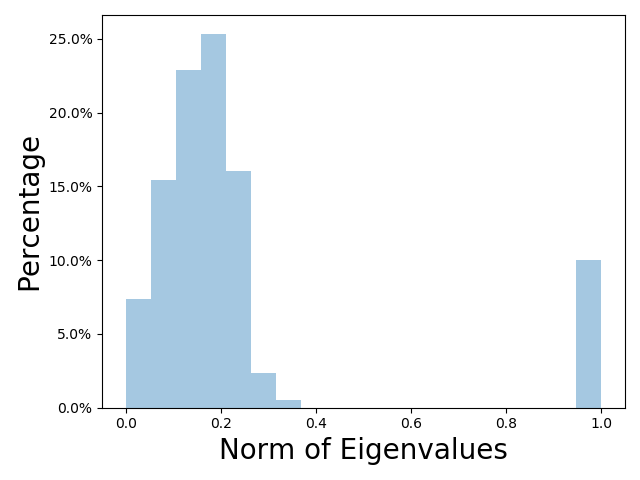}
        \caption{2 training points}
    \end{subfigure}%
    ~ 
    \begin{subfigure}{0.3\linewidth}
        \centering
        \includegraphics[height=1.2in]{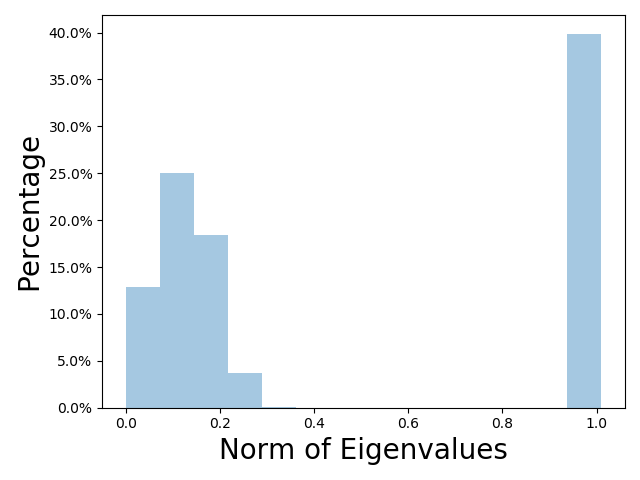}
        \caption{5 training points}
    \end{subfigure}
    ~
    \begin{subfigure}{0.3\linewidth}
        \centering
        \includegraphics[height=1.2in]{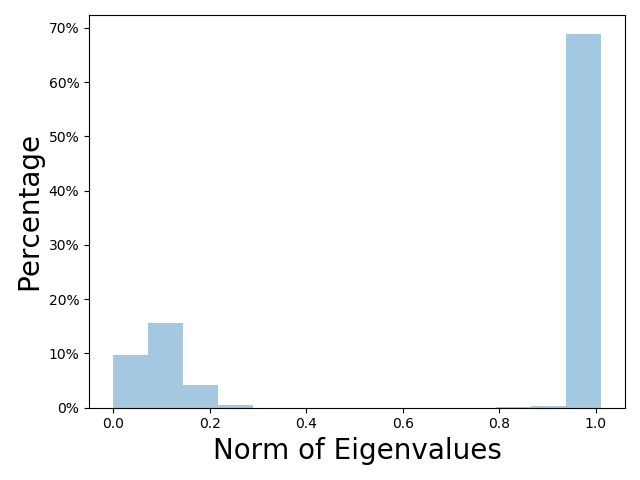}
        \caption{8 training points}
    \end{subfigure}
\caption{Eigenvalue distribution of 2-layer sigmoid network trained with input dimension 10}
\label{fig:2-layer-linear}
\end{figure*}

In this section, we first illustrate the eigenvalue distribution in the linear region. Here, we trained 2 layer $\mathrm{sigmoid}$ networks with input dimension $10$ and hidden size $1000$ for $2$, $5$ and $8$ training points. As suggested by Lemma~\ref{lemma:with-beta}, there should be $n-1$ eigenvalues with norm around $1$. This is supported by Figure~\ref{fig:2-layer-linear}, as there are $10\%$, $40\%$ and $70\%$ eigenvalues around that region. 
\label{append:linear}

\subsection{Basin of Attraction}
\label{append:basin}
We test basin of attraction by adding Gaussian noises to training examples and check if the modified examples can converge to the original ones via iterative maps under 50 iterations. The standard deviation of the Gaussian noise is called the noise radius. The network has 2 layers with hidden size $10000$ and input dimension $32$. Figure~\ref{fig:basin-all}
details experiments for $5$, $20$ and $40$ examples. Not surprisingly, the basin of attraction is larger when there are fewer training examples and larger input norms since a level of separation between data is required.

\begin{figure*}
    \centering
    \begin{subfigure}{0.3\linewidth}
        \centering
        \includegraphics[height=1.2in]{simulation/basin/32_10000_5_points.png}
        \caption{5 training points}
    \end{subfigure}%
    ~ 
    \begin{subfigure}{0.3\linewidth}
        \centering
        \includegraphics[height=1.2in]{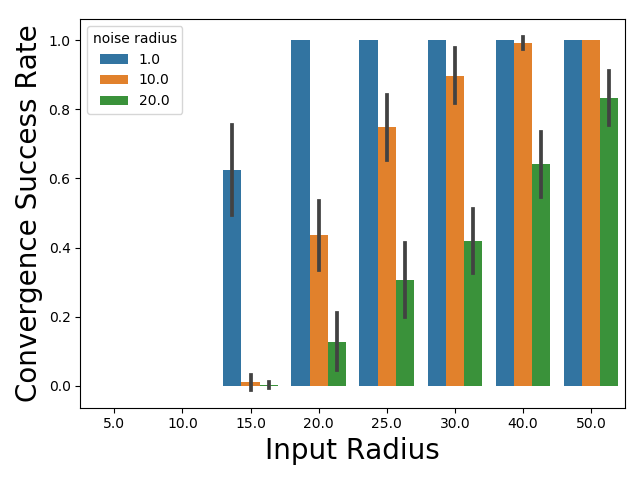}
        \caption{20 training points}
    \end{subfigure}
    ~
    \begin{subfigure}{0.3\linewidth}
        \centering
        \includegraphics[height=1.2in]{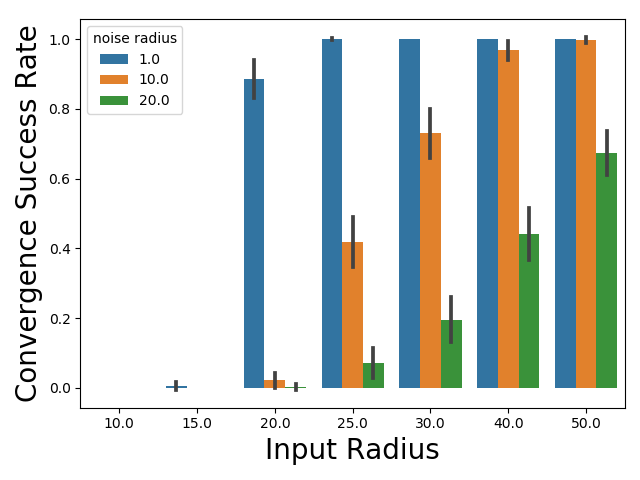}
        \caption{40 training points}
    \end{subfigure}
\caption{Convergence success rate vs input norm: random data with input dimension 32}
\label{fig:basin-all}
\end{figure*}

\begin{figure*}
    \centering
    \begin{subfigure}{0.4\linewidth}
        \centering
        \includegraphics[height=1.2in]{simulation/mnist_basin/784_10000_5_points.png}
        \caption{5 training points}
    \end{subfigure}%
    ~ 
    \begin{subfigure}{0.4\linewidth}
        \centering
        \includegraphics[height=1.2in]{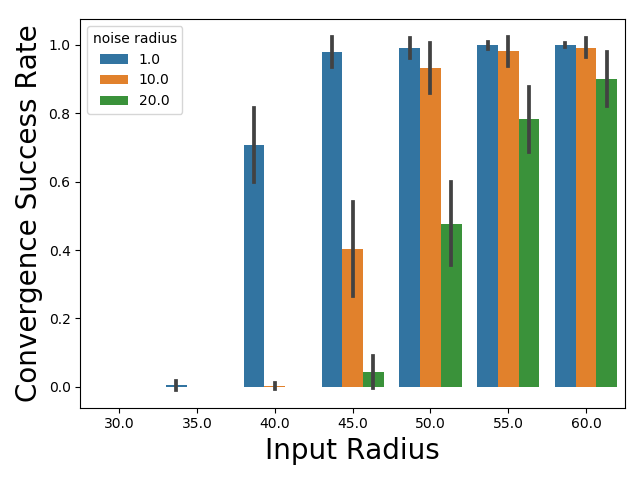}
        \caption{20 training points}
    \end{subfigure}
\caption{Convergence success rate vs input norm: MNIST dataset}
\label{fig:basin-all}
\end{figure*}

\begin{figure*}
    \centering
    \begin{subfigure}{0.3\linewidth}
        \centering
        \includegraphics[height=1.2in]{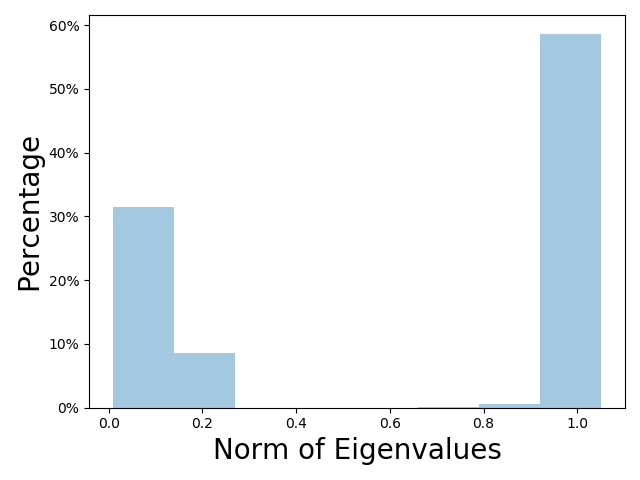}
        \caption{Radius: 1}
    \end{subfigure}%
    ~ 
    \begin{subfigure}{0.3\linewidth}
        \centering
        \includegraphics[height=1.2in]{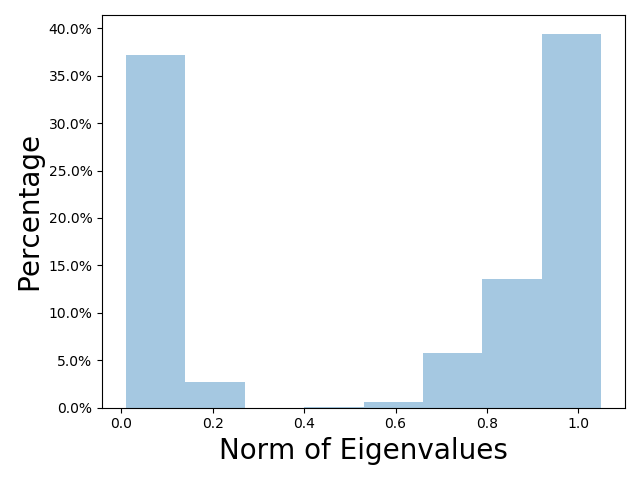}
        \caption{Radius: 5}
    \end{subfigure}%
    ~ 
    \begin{subfigure}{0.3\linewidth}
        \centering
        \includegraphics[height=1.2in]{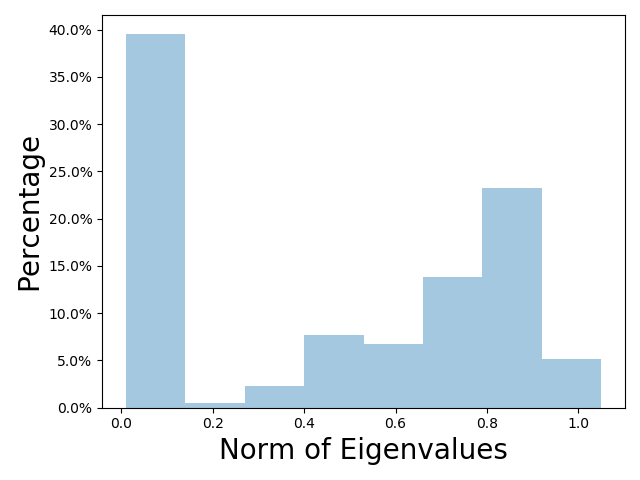}
        \caption{Radius: 10}
    \end{subfigure}
    
    \medskip
    
    \begin{subfigure}{0.3\linewidth}
    \centering
    \includegraphics[height=1.2in]{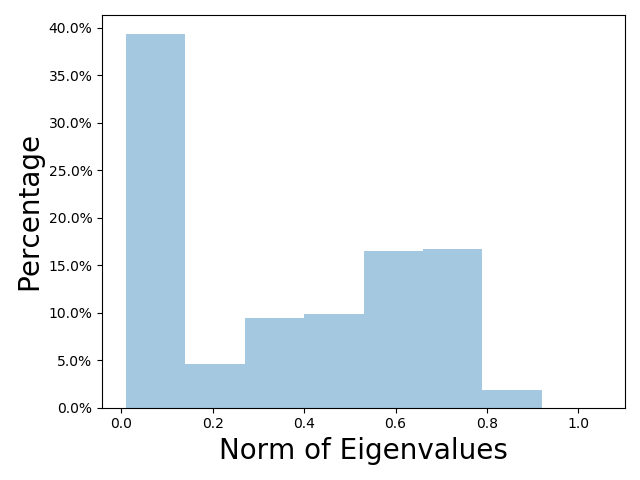}
    \caption{Radius: 15}
    \end{subfigure}%
    ~ 
    \begin{subfigure}{0.3\linewidth}
        \centering
        \includegraphics[height=1.2in]{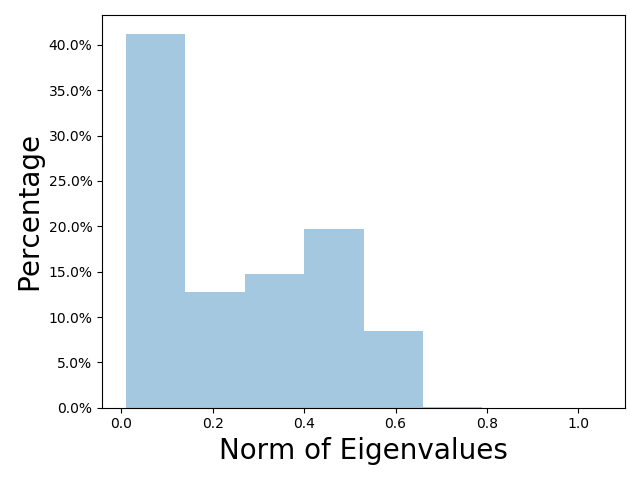}
        \caption{Radius: 20}
    \end{subfigure}
    \caption{Specturm Change for Sigmoid}
\label{fig:32-basin-sigmoid}
\end{figure*}

\begin{figure*}[b]
    \centering
    \begin{subfigure}{0.3\linewidth}
        \centering
        \includegraphics[height=1.2in]{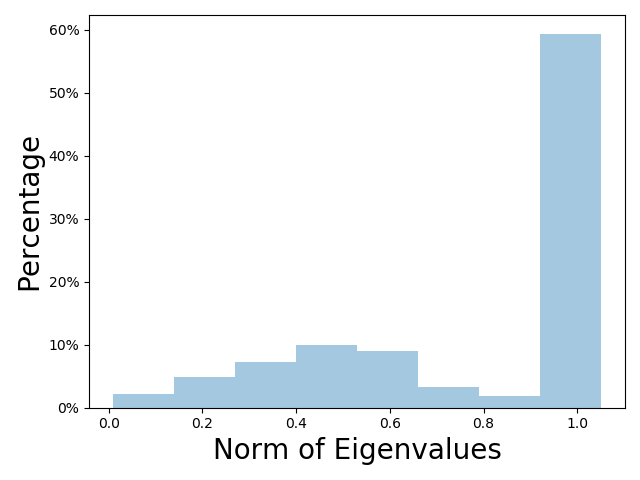}
        \caption{Radius: 1}
    \end{subfigure}%
    ~ 
    \begin{subfigure}{0.3\linewidth}
        \centering
        \includegraphics[height=1.2in]{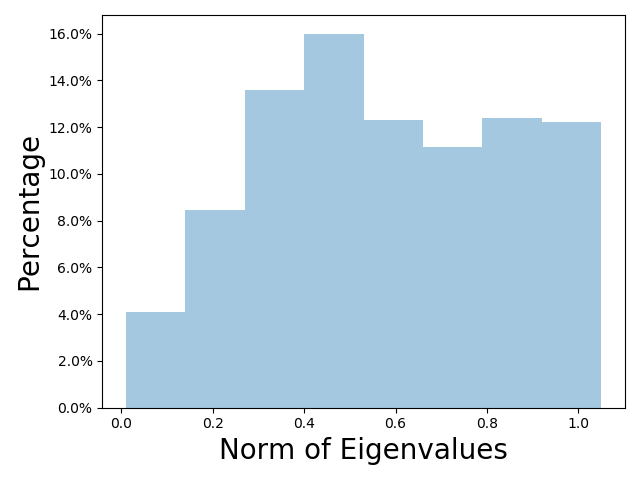}
        \caption{Radius: 5}
    \end{subfigure}%
    ~ 
    \begin{subfigure}{0.3\linewidth}
        \centering
        \includegraphics[height=1.2in]{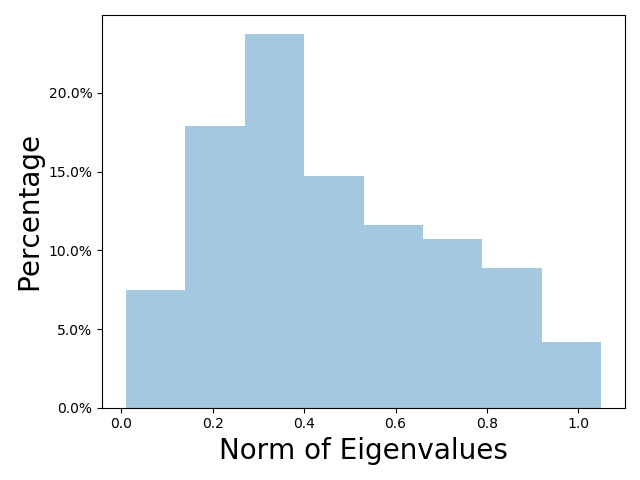}
        \caption{Radius: 10}
    \end{subfigure}
    
    \medskip
    
    \begin{subfigure}{0.3\linewidth}
    \centering
    \includegraphics[height=1.2in]{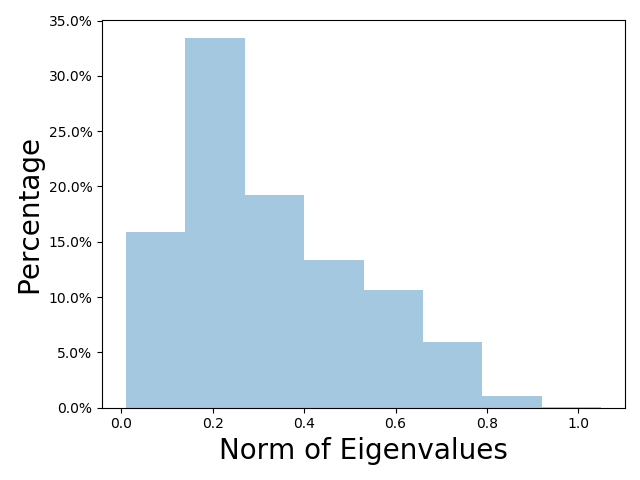}
    \caption{Radius: 15}
    \end{subfigure}%
    ~ 
    \begin{subfigure}{0.3\linewidth}
        \centering
        \includegraphics[height=1.2in]{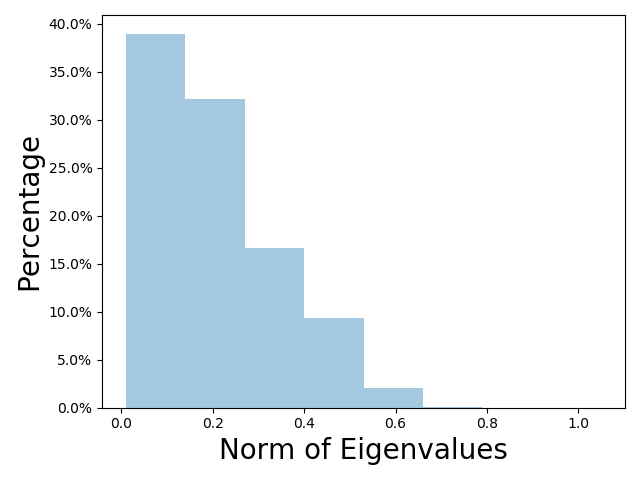}
        \caption{Radius: 20}
    \end{subfigure}
    \caption{Specturm Change for Erf}
\label{fig:32-basin-erf}
\end{figure*}

\subsection{Basin of Attraction on Mnist}
\label{append:basin-mnist}
We also test  basin of attraction experiments on MNIST dataset to check if we can recover real training examples. The images are prepossessed by subtracting means and rescaled to have different input norms for testing. Similar to the setting before, Figure~\ref{fig:basin--mnist-5} also shows that larger input norm gives greater basin of attraction for $5$ and $20$ examples. Notice that because MNIST images have large input dimension, they need larger radius to move out of the linear region.

\subsection{Sigmoidal Activations}
\label{append:act-eigen-histo}

\begin{figure*}
    \centering
    \begin{subfigure}{0.3\linewidth}
        \centering
        \includegraphics[height=1.2in]{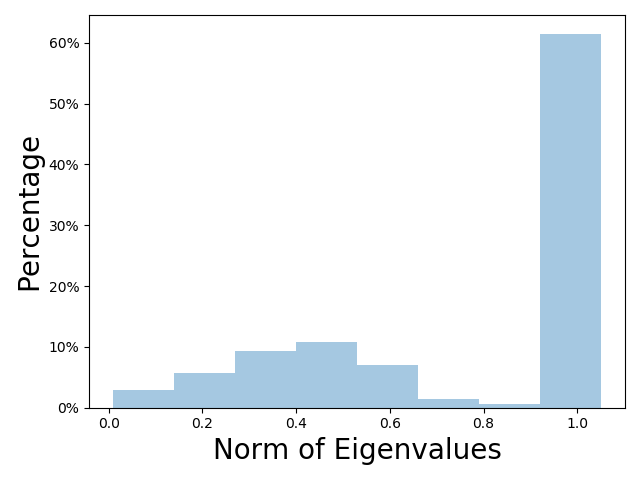}
        \caption{Radius: 1}
    \end{subfigure}%
    ~ 
    \begin{subfigure}{0.3\linewidth}
        \centering
        \includegraphics[height=1.2in]{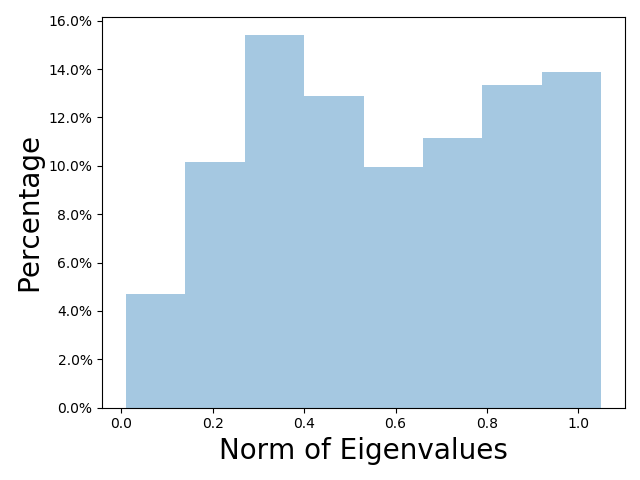}
        \caption{Radius: 5}
    \end{subfigure}%
    ~ 
    \begin{subfigure}{0.3\linewidth}
        \centering
        \includegraphics[height=1.2in]{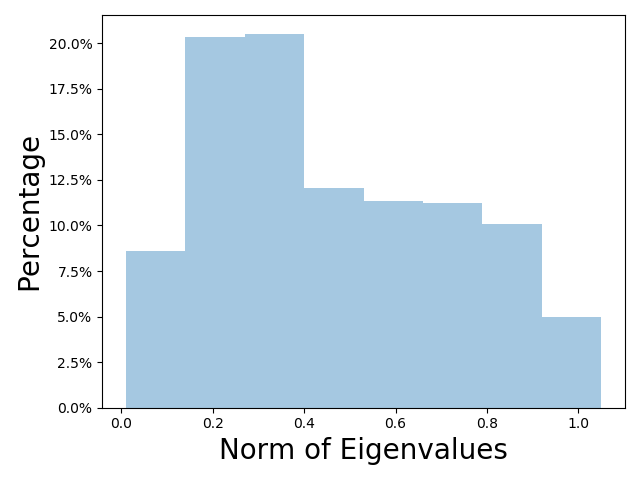}
        \caption{Radius: 10}
    \end{subfigure}
    
    \medskip
    
    \begin{subfigure}{0.3\linewidth}
    \centering
    \includegraphics[height=1.2in]{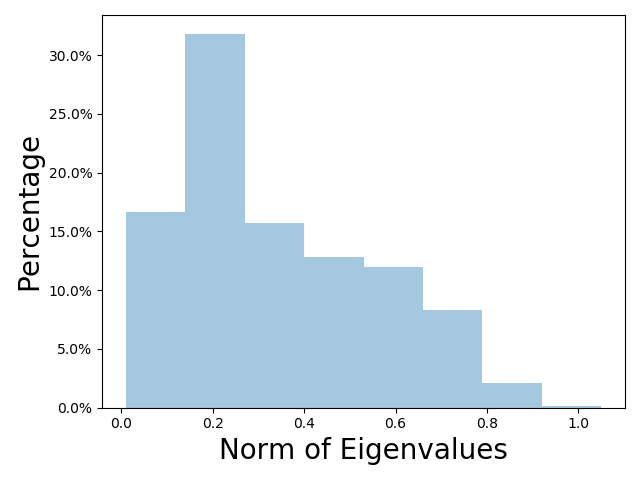}
    \caption{Radius: 15}
    \end{subfigure}%
    ~ 
    \begin{subfigure}{0.3\linewidth}
        \centering
        \includegraphics[height=1.2in]{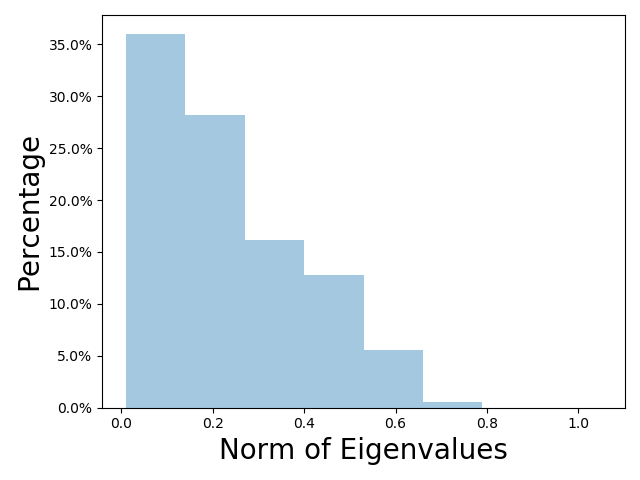}
        \caption{Radius: 20}
    \end{subfigure}
    \caption{Specturm Change for Sigmoid}
\label{fig:32-basin-tanh}
\end{figure*}

Finally, we show that our results can be extended to different sigmoidal activation functions as well. We chose 2 layer network with hidden size $10000$, input dimension $32$ and $20$ training examples. As before, only settings that can let network converges to training loss below $10^{-7}$ are included. Figure~\ref{fig:act} clearly suggests all the activation functions share similar curves. Notice that both $\mathrm{tanh}$ and $\mathrm{erf}$ have large eigenvalue when $r$ is small. This is not a contradiction to our Lemma~\ref{lemma:no-beta} as their $\alpha = \dot{\sigma}(0)$ is too large to satisfy the conditions in Lemma~\ref{lemma:no-beta}. The histogram of eigenvalue norm changes for those activation is shown in Figure~\ref{fig:32-basin-sigmoid}, Figure~\ref{fig:32-basin-erf}, Figure~\ref{fig:32-basin-tanh}. It is clear that they all follow the same pattern. 

\end{document}